\documentclass[journal]{IEEEtran}

\usepackage{amsmath,graphicx}

\usepackage{hyperref}
\usepackage[utf8]{inputenc}
\usepackage{tabularx}
\usepackage{setspace}
\usepackage{graphicx}
\usepackage{booktabs}
\usepackage{amsthm}
\usepackage{color}
\usepackage{multirow}
\usepackage{comment} 

\usepackage{url,cite} 
\usepackage{caption}
\usepackage{verbatim} 
\usepackage{amssymb}
\usepackage{algorithm}
\usepackage{algorithmic}
\usepackage{tikz}
\usepackage{footnote}
\usepackage{url}
\usepackage{bm}
\usepackage{arydshln}

\begin{document}


\title{Backpropagation with $N$-D Vector-Valued Neurons Using Arbitrary Bilinear Products}

\author{Zhe-Cheng~Fan,~
        Tak-Shing~T.~Chan,~\IEEEmembership{Member,~IEEE,}
        Yi-Hsuan~Yang,~\IEEEmembership{Senior~Member,}
        and~Jyh-Shing~R.~Jang,~\IEEEmembership{Member,~IEEE}

}

\markboth{Journal of \LaTeX\ Class Files,~Vol.~14, No.~8, August~2015}%
{Shell \MakeLowercase{\textit{et al.}}: Bare Demo of IEEEtran.cls for IEEE Journals}

\maketitle

\begin{abstract}

Vector-valued neural learning has emerged as a promising direction in deep learning recently. Traditionally, training data for neural networks (NNs) are formulated as a vector of scalars; however, its performance may not be optimal since associations among adjacent scalars are not modeled. In this paper, we propose a new vector neural architecture called the Arbitrary BIlinear Product Neural Network (ABIPNN), which processes information as vectors in each neuron, and the feedforward projections are defined using arbitrary bilinear products. Such bilinear products can include circular convolution, seven-dimensional vector product, skew circular convolution, reversed-time circular convolution, or other new products not seen in previous work. 
As a proof-of-concept, we apply our proposed network to multispectral image denoising and singing voice separation. Experimental results show that ABIPNN gains substantial improvements when compared to conventional NNs, suggesting that associations are learned during training. 

\end{abstract}


\IEEEpeerreviewmaketitle

\section{Introduction}

\IEEEPARstart{V}{ector}-valued neurons have received much attention lately in different scientific fields, such as communication systems, biological processing, image processing, and audio signal processing \cite{goh09complex, nitta09complex, hirose13springer, hirose13wiley, bengio18iclr}. Each training sample of these applications can be represented as a multidimensional vector, which can be processed directly by vector-valued neurons. These aforementioned works have shown that vector-valued neurons have good performance in learning, association, and generalization. In the meantime, there are more and more datasets \cite{yasuma10itip,koelstra12tac,vincent15chime} that provide multidimensional data suitable for vector-valued neural learning.

In real-valued neural network (NN) learning with multidimensional data, the input is concatenated from a set of vectors and reformulated as a one-dimensional vector. A neuron takes only one real value as its input and a network is configured to use as many neurons as the dimension of the one-dimensional vector. But this configuration may not achieve satisfactory performance for multidimensional problems since associations within each vector are not learned. Therefore, multidimensional vector neurons have received some attention in the literature and have been proposed to address the associations among different dimensions. A vector-valued neuron accepts and represents information as a vector and elements in each vector are processed together as a single unit. 

There are several approaches to extend real-valued neurons to vector-valued ones. 
Two-dimensional complex-valued NNs\cite{nitta97nn} are proven to have orthogonal decision boundaries\cite{nitta00npl,nitta04nc} and the ability to solve exclusive-or (XOR) problem\cite{nitta03nn} using only a single neuron. 
Another extension to two dimensions is the hyperbolic neural network \cite{buchholz00ijcnn}, in which all parameters are hyperbolic numbers. Decision boundaries of hyperbolic neurons are investigated in \cite{nitta08ijcnn}. An alternative hyperbolic backpropagation algorithm \cite{nitta17tnnls} is developed using Wirtinger calculus.
Three-dimensional neurons have been proposed in two ways. The first one is based on the vector product\cite{bkt15hdn} \cite{nitta93ijcnn}, in which inputs, outputs, weights and biases are all three-dimensional vectors. The second one \cite{nitta06nipr} is similar to the first but the weights are now three-dimensional orthogonal matrices. 
Four-dimensional hypercomplex-valued neurons \cite{arena94iscas, arena98Springer, nitta95icnn, bkt15hdn} are proposed by using the quaternion algebra.
Eight-dimensional octonion-valued neurons\cite{popa16icann} represent a generalization of quaternion NNs. More recently, deep complex networks \cite{bengio18iclr} are introduced using complex convolution and complex batch normalization.

\begin{figure*}
\begin{center}
\includegraphics[trim=4.2cm 3cm 4.0cm 1cm, clip, scale=0.40]{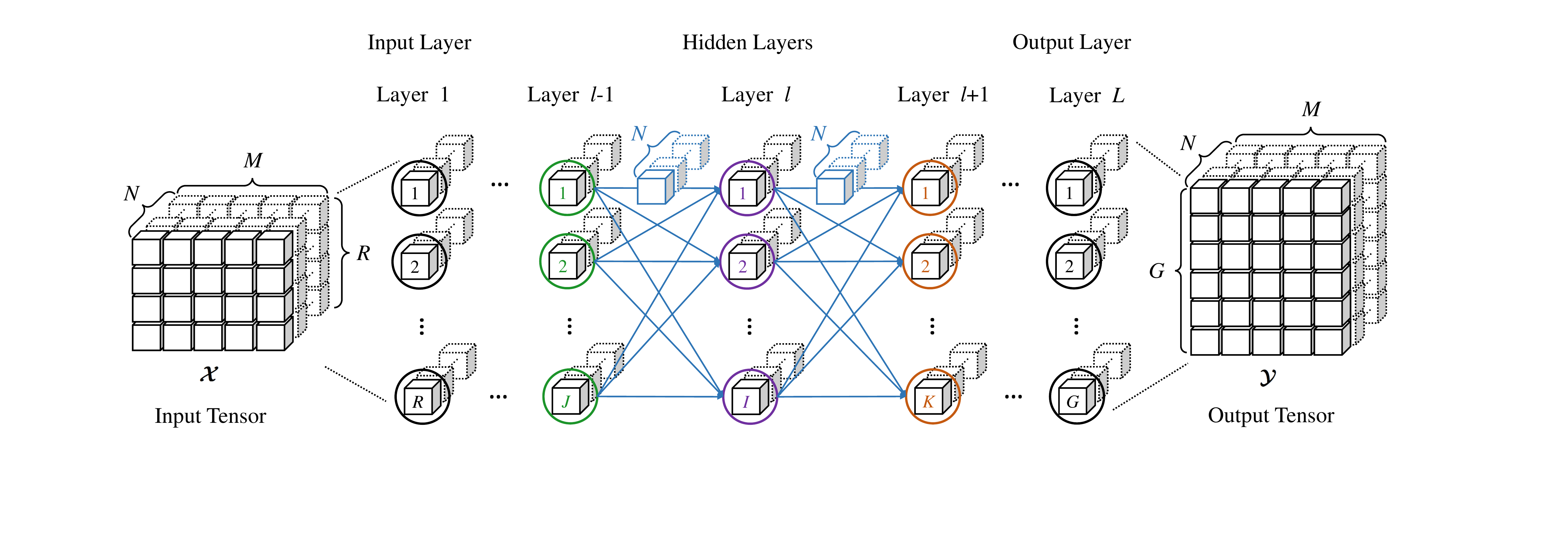}
\caption{Architecture of ABIPNN (best viewed in color), where the inputs, outputs, weights, and biases are all $N$-dimensional vectors. Cubes in the architecture represent scalars. Blue lines between hidden layers represent weights. $M$ represents the number of training data. $R$ and $G$ are the dimensions of the input and output layers, respectively. $J$, $I$, and $K$ are the dimensions of the hidden layers. $\bm{\mathcal{X}}$ and $\bm{\mathcal{Y}}$ represents the input and output tensors, respectively. When $N$ equals one, each neuron represents a scalar, depicted as a cube with solid lines. This architecture can be viewed as a conventional NN. When $N>1$, each neuron represents a vector, depicted as the concatenation of cubes connected by the dotted lines.}
\label{fig:vnn_arch}
\end{center}
\end{figure*}

However, the dimensionality of a vector-valued neuron should not be constrained to a particular number. For instance, in the task of multispectral image denoising \cite{zhang16ijcai}, different bands of images are stacked into a tensor. Nonlinear mapping between the noisy tensor and the clean tensor is then performed. For singing voice separation\cite{fan17dlm}, the data structure of a temporal-frequency matrix input is flattened into a high-dimensional vector in the traditional way then reshaped as a matrix so that each neuron represents and receives a vector. For EEG-based emotion recognition \cite{koelstra12tac,jirayucharoensak14tswj}, the human brain wave is filtered into five main frequency bands. Given the above observations, it seems important that the dimensionality should be arbitrary $N$. 
Thus, a good extension of the real-valued neuron is the $N$-dimensional real-valued neuron \cite{bkt15hdn,nitta07ijcai,nitta13wiley}, which was proposed to have $N$-dimensional vector inputs and outputs, but $N$-dimensional orthogonal matrix weights. Nevertheless, we observe that this formulation does not address the case of multiple neurons.
Other architectures have also been advanced to extend real-valued neurons. For example, 
Clifford algebra has been employed to build vector-valued NNs with dimensionality $2^N$ \cite{pearson92nn,pearson94nn,bayro01geometric,Sommer01camp}; however, we note that many datasets do not have power-of-two dimensionalities. 
Matrix-valued NNs\cite{popa15mvnn} have been proposed in which the inputs, outputs, weights, and biases are $N\times N$ square matrices. But it is not always easy to formulate the inputs and outputs like this.
Finally, for multiway classification, the tensor-factorized NN \cite{chien17tnnls} integrates Tucker decomposition with neural learning, though its efficiency in regression problems is yet unproven.

In light of the above observations, we propose a new vector-valued NN architecture consisting of $N$-dimensional vector-valued neurons, where $N$ is an arbitrary positive integer (as shown in Fig.~\ref{fig:vnn_arch}). For each neuron, the inputs, outputs, weights, and biases are all $N$-dimensional vectors. Our key observation is that the products used by the aforementioned vector-valued NNs (such as the vector product~\cite{nitta93ijcnn} and quaternion multiplication~\cite{arena94iscas}) are bilinear products (defined in Section~\ref{sec:vector_neural_learning}). This prompts us to propose a general form of bilinear neurons to model the associations between the vector elements. We call it the Arbitrary BIlinear Product Neural Network (ABIPNN). When $N$ equals one, each neuron represents a scalar and the architecture performs matrix multiplications like a conventional deep neural network (DNN). When $N$ is larger than one, each neuron represents a vector and the architecture uses bilinear products for multiplications. In this way, the proposed architecture not only allows for using vectors of arbitrary dimensionality in vector-valued NNs, but also all the vector-valued products that are bilinear.

The remainder of this paper is organized as follows. Section \ref{sec:vector_neural_learning} derives the feedforward and backpropagation processes using the proposed bilinear neurons. Section \ref{sec:experiments} compares the performance between ABIPNN and conventional NNs in singing voice separation and multispectral image denoising. We conclude the paper in Section \ref{sec:conclusion}.

\begin{figure}
\begin{center}
\includegraphics[trim=8cm 2cm 7cm 1.5cm, clip, scale=0.42]{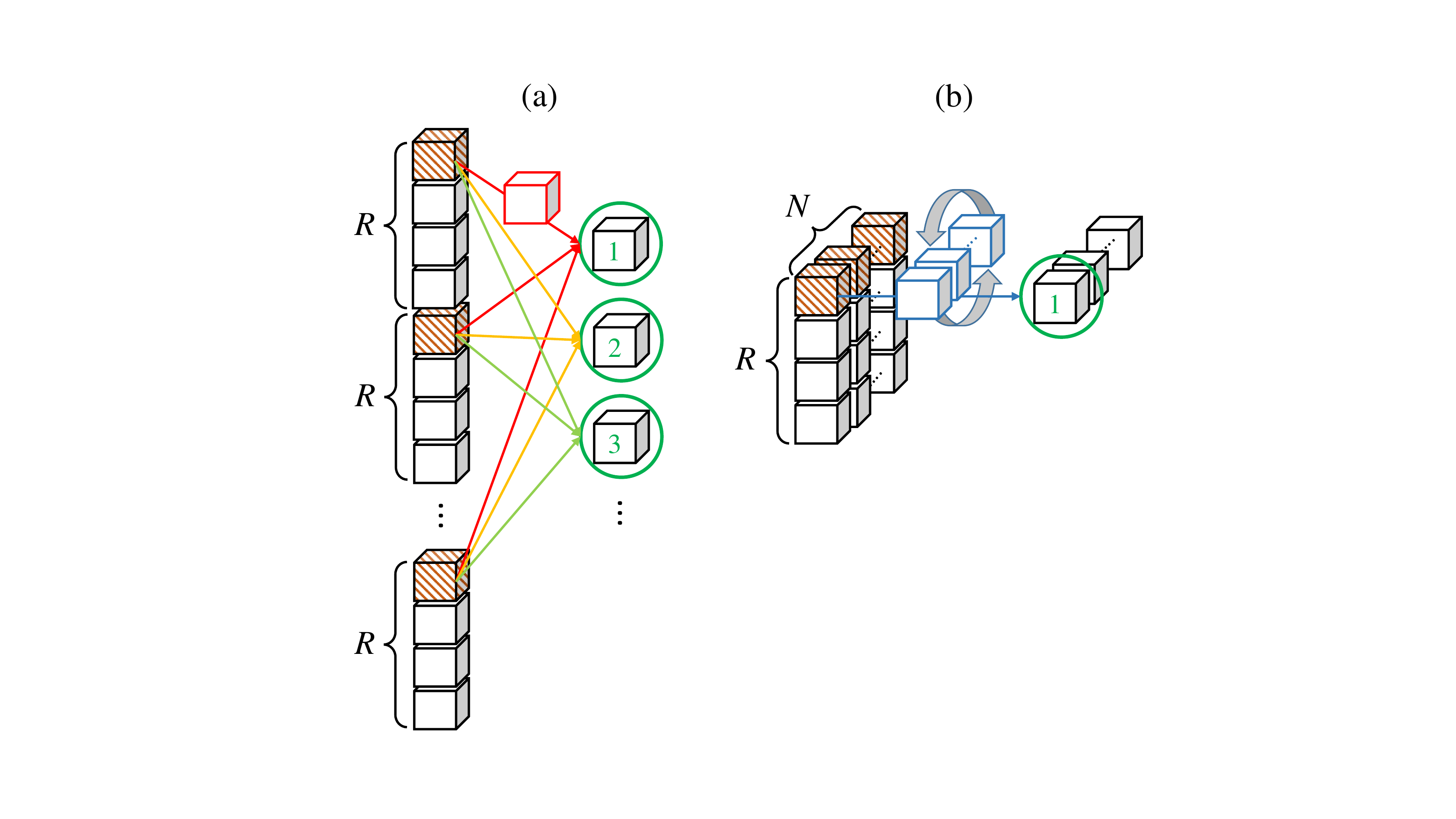}
\caption{Illustration of the operations employed by the (a) conventional NNs and (b) ABIPNNs; best viewed in color. In (a), vectors are concatenated together as a multidimensional input. Each weight stands for a  scalar, depicted as a cube. Take the first entry of each vector for example, the associations between the brown cubes are not learned since each hidden neuron receives information by using its group of weights, which are depicted in the same color. Hidden neurons do not share identical weights when performing learning. In (b), vectors are stacked into a $R\times 1\times N$ tensor (or lateral matrix) and brown cubes are concatenated together. Each neuron represents a vector and receives information by using the weight vector, depicted as blue cubes. For example, when we leverage circular convolution as the bilinear product in an ABIPNN, the weight vector can be deemed as a linear kernel mask which captures the associations between brown cubes by rotating itself in the learning process.}
\label{fig:vnn_explain}
\end{center}
\end{figure}

\section{Vector neural learning}
\label{sec:vector_neural_learning}

The idea of ABIPNN is to extend the data type of each neuron from scalars to vectors by replacing multiplications with arbitrary bilinear products. Such products can be used to learn the associations between vector elements in the input (see Fig.~\ref{fig:vnn_explain}). 
In the following, tensors are represented by bold uppercase calligraphic letters, matrices by bold uppercase letters, and vectors by bold lowercase letters. Matrix slices of tensors are represented as matrices and vector slices of matrices are represented as vectors. The notational conventions used in this paper are summarized in Table \ref{tab:convention_table}. 

\subsection{Generalizing N-D Vector Neurons with Bilinear Products}
To begin with, we identify a generalization of all the products used in previous vector-valued NN literature, including the vector product~\cite{nitta93ijcnn}, quaternion multiplication~\cite{arena94iscas}, octonion multiplication~\cite{popa16icann}, and so on. We have been able to confirm that all of these products are bilinear (defined below). Motivated by this important observation, we will extend the vector-valued NN \cite{nitta07ijcai} to use any kind of bilinear product between two $N$-dimensional vectors.

Assume an $N$-dimensional space with standard basis $\left\{\mathbf{e}_1,~\mathbf{e}_2,~\ldots,~\mathbf{e}_N\right\}$ as column vectors. Let $\mathbf{p}=\sum_{n=1}^N p_n\mathbf{e}_n=\left[p_1,~p_2,~\ldots,~p_N\right]^T$ and $\mathbf{q}=\sum_{n=1}^N q_n\mathbf{e}_n=\left[q_1,~q_2,~\ldots,~q_N\right]^T$ be two vectors in this space. A product $\bullet\colon \mathbb{R}^N\times \mathbb{R}^N\to \mathbb{R}^N$ is bilinear if and only if $\mathbf{p}\bullet \mathbf{q}$ is linear when we hold one of $\mathbf{p}$ or $\mathbf{q}$ fixed. If $\bullet$ is bilinear, we have:
\begin{equation}
\mathbf{p}\bullet \mathbf{q}=\mathbf{p}\bullet\left(\sum_{n=1}^N q_n\mathbf{e}_n\right)\textrm{,}
\end{equation}
where we have expanded the second term. Due to bilinearity,
\begin{equation}
\label{eq:biliner_product_second_term}
\begin{split}
\mathbf{p}\bullet \mathbf{q}&=\sum_{n=1}^N \left(\mathbf{p}\bullet \mathbf{e}_n\right)q_n\\
&=\left[\mathbf{p}\bullet 
\begin{bmatrix}
1 \\
0 \\
\vdots \\
0 
\end{bmatrix},
~\mathbf{p}\bullet 
\begin{bmatrix}
0 \\
1 \\
\vdots \\
0
\end{bmatrix},
~\ldots,~\mathbf{p}\bullet 
\begin{bmatrix}
0 \\
0 \\
\vdots \\
1 
\end{bmatrix}
\right]
\begin{bmatrix}
q_1 \\
q_2 \\
\vdots \\
q_N 
\end{bmatrix}\\
&=\left[\mathbf{p}\bullet \mathbf{e}_1,~\mathbf{p}\bullet \mathbf{e}_2,~\ldots,~\mathbf{p}\bullet \mathbf{e}_N\right]\mathbf{q}.
\end{split}
\end{equation}
If we let $\left[\mathbf{p}\right]_\bullet=\left[\mathbf{p}\bullet \mathbf{e}_1,~\mathbf{p}\bullet \mathbf{e}_2,~\ldots,~\mathbf{p}\bullet \mathbf{e}_N\right]$, then:
\begin{equation}
\label{eq:biliner_product_second_term_result}
\mathbf{p}\bullet \mathbf{q}=\left[\mathbf{p}\right]_\bullet \mathbf{q}\textrm{.}
\end{equation}
Similarly, if we expand the first term, we can write:
\begin{equation}
\label{eq:bilinear_product_first_term}
\begin{split}
\mathbf{p}\bullet\mathbf{q}
&=\left(\sum_{n=1}^{N}p_n\mathbf{e}_n\right)\bullet\mathbf{q} =\sum_{n=1}^{N}p_n\left(\mathbf{e}_{n}\bullet\mathbf{q}\right)\\
&=\left[\mathbf{e}_1\bullet\mathbf{q},~\mathbf{e}_2\bullet\mathbf{q},~\dots,~ \mathbf{e}_N\bullet\mathbf{q}\right]\mathbf{p}\textrm{.}
\end{split}
\end{equation}
Let $\left[\mathbf{q}\right]_\bullet^\dagger=\left[\mathbf{e}_1\bullet \mathbf{q},~\mathbf{e}_2\bullet \mathbf{q},~\ldots,~\mathbf{e}_N\bullet \mathbf{q}\right]$, we also have:
\begin{equation}
\label{eq:bilinear_product_first_term_result}
\mathbf{p}\bullet\mathbf{q}=\left[\mathbf{q}\right]_{\bullet}^{\dagger}\mathbf{p}.
\end{equation}
Eqs.~\ref{eq:biliner_product_second_term_result} and \ref{eq:bilinear_product_first_term_result} are the two possible matrix representations of the bilinear product $\bullet$. Here we call $\left[\mathbf{p}\right]_\bullet$ its matrix representation and $\left[\mathbf{q}\right]_\bullet^\dagger$ its transmuted representation (the term \emph{transmuted} is originally used for quaternions~\cite{ell07arxiv} but here we extend it for general bilinear products). In what follows, the feedforward and backpropagation processes will be described by using the above notations for bilinear products and their representations. For convenience, in the rest of this paper, if $\mathbf{p}$ or $\mathbf{q}$ are not column vectors, then $\mathbf{p}\bullet\mathbf{q}$ will implicitly reshape them so that the above equations make sense.

\begin{table}[t]
\renewcommand{\arraystretch}{1.2}
\caption{\textsc{Notational Conventions}}
\label{tab:convention_table}
\centering  
\begin{tabular}{|l l|}  
\hline
$\bm{\mathcal{X}}$                                    & Input tensor  \\
$\mathbf{X}_{m}$=$\bm{\mathcal{X}}_{:m:}$             & Lateral slice from input tensor $\bm{\mathcal{X}}$ \\
$\mathbf{x}$                                          & Vector from input tensor $\bm{\mathcal{X}}$  \\
$\bm{\mathcal{Y}}$                                    & Output tensor  \\
$\mathbf{Y}_{m}$=$\bm{\mathcal{Y}}_{:m:}$             & Lateral slice from output tensor $\bm{\mathcal{X}}$ \\
$\bm{\mathcal{W}}^{l}$                                & Weight tensor in layer $l$        \\ 
$\mathbf{w}_{ij}^{l}=\bm{\mathcal{W}}_{ij:}^{l}$      & Vector from tensor $\bm{\mathcal{W}}^{l}$  \\
$\bm{\mathcal{B}}_{}^{l}$                             & Bias tensor in layer $l$        \\ 
$\mathbf{b}_{i}^{l}=\bm{\mathcal{B}}_{im:}^{l}$       & Vector from tensor $\bm{\mathcal{B}}^{l}$  \\
$\bm{\mathcal{Z}}^{l}$                                & Input hidden tensor in layer $l$        \\ 
$\mathbf{z}_{i}^{l}=\bm{\mathcal{Z}}_{im:}^{l}$       & Vector from tensor $\bm{\mathcal{Z}}^{l}$  \\
$\bm{\mathcal{A}}^{l}$                                & Output hidden tensor in layer $l$        \\ 
$\mathbf{a}_{i}^{l}=\bm{\mathcal{A}}_{im:}^{l}$       & Vector from tensor $\bm{\mathcal{A}}^{l}$  \\ 
$\bm{\mathcal{D}}^{l}$                                & Local gradient tensor in layer $l$        \\ 
$\mathbf{d}_{i}^{l}=\bm{\mathcal{D}}_{km:}^{l}$       & Vector from tensor $\bm{\mathcal{D}}^{l}$  \\ 
$\bullet\colon \mathbb{R}^N\times \mathbb{R}^N\to \mathbb{R}^N$ & Arbitrary bilinear product \\
$\left[\cdot\right]_\bullet$                          & Matrix representation of $\bullet$ (Eq.~\ref{eq:biliner_product_second_term_result})\\
$\left[\cdot\right]_\bullet^\dagger$                  & Transmuted representation of $\bullet$ (Eq.~\ref{eq:bilinear_product_first_term_result})\\
\hline
\end{tabular}
\end{table}

\subsection{Feedforward Process in a Bilinear Product Neuron}
\label{subsec:feedforward}
As before, we assume the inputs, outputs, weights, and biases are $N$-dimensional vectors. Suppose we have an $L$-layer vector-valued NN. For hidden layer $l$ ($1 \leq l \leq L$), the input vector $\mathbf{z}_i^l$ of the hidden neuron $i$ is defined as:
\begin{equation}
\label{eq:feedforward}
\mathbf{z}_i^l=\sum_{j=1}^{J}\mathbf{w}_{ij}^{l}\bullet\mathbf{a}_j^{l-1}+\mathbf{b}_{i}^{l}~\textrm{,}
\end{equation}
where $\bullet$ denotes an arbitrary bilinear product, $\mathbf{w}_{ij}^l$ stands for the weight vector connecting a neuron $j$ ($1 \leq j \leq J$) in layer $l-1$ to a neuron $i$ ($1 \leq i \leq I$) in layer $l$, and
$\mathbf{w}_{ij}^l=\sum_{n=1}^N w_{ijn}^l\mathbf{e}_n=\left[w_{ij1}^l\textrm{,}~w_{ij2}^l\textrm{,}\dots\textrm{,}w_{ijN}^l\right]^T\in \mathbb{R}^N$. 
The output vector of neuron $j$ in layer $l-1$ is $\mathbf{a}_{j}^{l-1}=\sum_{n=1}^N a_{jn}^{l-1}\mathbf{e}_n=\left[a_{j1}^{l-1}\textrm{,}~a_{j2}^{l-1}\textrm{,}\dots\textrm{,}~a_{jN}^{l-1}\right]^T\in \mathbb{R}^N$,
the bias vector of neuron $i$ in layer $l$ is $\mathbf{b}_{i}^l=\sum_{n=1}^N b_{in}^{l}\mathbf{e}_n=\left[b_{i1}^l\textrm{,}~b_{i2}^l\textrm{,}\dots\textrm{,}~b_{iN}^l\right]^T \in \mathbb{R}^N$. 
As a result, $\mathbf{z}_i^l=\sum_{n=1}^N z_{in}^{l}\mathbf{e}_n=\left[z_{i1}^l\textrm{,}~z_{i2}^l\textrm{,}\dots\textrm{,}~z_{iN}^l\right]^T \in \mathbb{R}^N$. 
When $l$ equals $1$, the vector $\mathbf{z}$ is simply the input vector $\mathbf{x}$. After $\mathbf{z}_i^l$ is calculated, the output vector $\mathbf{a}_{i}^{l}$ of the hidden neuron $i$ is as follows:
\renewcommand{\arraystretch}{1.2}
\begin{equation}
\label{eq:feedforeard_at_layer_l}
\mathbf{a}_i^l=
\begin{bmatrix}
a_{i1}^l \\
a_{i2}^l \\
\vdots \\
a_{iN}^l
\end{bmatrix}
=\phi(\mathbf{z}_i^l)=
\begin{bmatrix}
\phi(z_{i1}^l) \\
\phi(z_{i2}^l) \\
\vdots           \\
\phi(z_{iN}^l)
\end{bmatrix} \textrm{,}
\end{equation}
where $\phi$ could be any differentiable activation function. Then the output vector of neuron $g$ in output layer $L$ is defined as:
\renewcommand{\arraystretch}{1.2}
\begin{equation}
\label{eq:feedforeard_at_layer_L}
\mathbf{y}_g^L=
\begin{bmatrix}
y_{g1}^{L} \\
y_{g2}^{L} \\
\vdots \\
y_{gN}^{L}
\end{bmatrix}
=\phi(\mathbf{z}_g^L)=
\begin{bmatrix}
\phi(z_{g1}^{L}) \\
\phi(z_{g2}^{L}) \\
\vdots           \\
\phi(z_{gN}^{L})
\end{bmatrix} \textrm{,}
\end{equation}
where $\mathbf{z}_g^L$ is the input vector and $\mathbf{y}_g^L$ is the output vector of a neuron $g$. 
The objective of the training process is to estimate the parameters that minimizes the cost function, defined as:  
\begin{equation}
\label{eq:cost_function}
C(\bm{\Theta})=\sum_{m=1}^M loss(\mathbf{Y}_m, f(\mathbf{X}_m;\bm{\Theta}))~\textrm{,}
\end{equation}
where $M$ stands for number of training data and $\mathbf{\Theta}$ is the set of all training parameters (weights and biases), and $\mathbf{Y}_m$ is the training label related to the input $\bm{X}_m$. The output $f(\mathbf{X}_m;\mathbf{\Theta})$ is the predicted version of $\mathbf{Y}_m$, made up of $\mathbf{y}_g^L$ mentioned in Eq.~\ref{eq:cost_function}. The $loss(\mathbf{Y}_m, f(\mathbf{X}_m;\bm{\Theta}))$
function measures the difference between the predicted results and the training labels.

\subsection{Backpropagation Algorithm in a Bilinear Product Neuron}
\label{subsec:backpropagate}
The bilinear product is also employed in the process of backpropagation learning. 
Before we present the error backpropagation process (Algorithm~\ref{alg:ABIPalg}), we need to first derive:
\begin{enumerate}
\item The gradients of the biases $\mathbf{b}_{i}^{l}$.
\item The gradients of the weights $\mathbf{w}_{ij}^{l}$.
\item The backpropagation of local gradients $\mathbf{d}_{k}^{l+1}\rightarrow\mathbf{d}_{i}^{l}$.
\end{enumerate}
Following Eq.~\ref{eq:biliner_product_second_term_result}, the bilinear product can be viewed as matrix-vector multiplication. Hence, Eq.~\ref{eq:feedforward} can be rewritten as
\renewcommand{\arraystretch}{1.2}
\begin{equation}
\begin{split}
\label{eq:feedforward_extend_part1}
\mathbf{z}_i^l
&=\sum_{j=1}^{J}\mathbf{w}_{ij}^{l}\bullet\mathbf{a}_{j}^{l-1}+\mathbf{b}_{i}^{l} =\sum_{j=1}^{J}\left[\mathbf{w}_{ij}^{l}\right]_{\bullet}\mathbf{a}_{j}^{l-1}+\mathbf{b}_{i}^{l}\\
&=\sum_{j=1}^{J}\left[
\mathbf{w}_{ij}^l\bullet\mathbf{e}_1\textrm{,}~
\mathbf{w}_{ij}^l\bullet\mathbf{e}_2\textrm{,}~
\dots \textrm{,}
\mathbf{w}_{ij}^l\bullet\mathbf{e}_N\right]\mathbf{a}_{j}^{l-1}+\mathbf{b}_{i}^{l}\\
&=\sum_{j=1}^{J}\left[
a_{j1}^{l-1}\mathbf{w}_{ij}^l\bullet\mathbf{e}_1\textrm{,}~
\dots \textrm{,}
a_{jN}^{l-1}\mathbf{w}_{ij}^l\bullet\mathbf{e}_N\right] +\mathbf{b}_{i}^{l}\textrm{.}\\
\end{split}
\end{equation}
Here, $\mathbf{w}_{ij}^{l}$ can be formulated as the summation of scalar-vector multiplications:
\begin{equation}
\begin{split}
\label{eq:feedforward_extend_part2}
\mathbf{z}_i^l=
&\sum_{j=1}^{J}\left(
a_{j1}^{l-1}
\left(
w_{ij1}^l \mathbf{e}_1+
\dots+w_{ijN}^l \mathbf{e}_N \right)\right. \bullet\mathbf{e}_1  \\
&\quad\thinspace+
a_{j2}^{l-1}
\left(
w_{ij1}^l \mathbf{e}_1+
\dots+w_{ijN}^l \mathbf{e}_N \right) \bullet\mathbf{e}_2 + \dots\\
&\quad\thinspace+
a_{jN}^{l-1}\left.
\left(
w_{ij1}^l \mathbf{e}_1+
\dots+w_{ijN}^l \mathbf{e}_N \right) \bullet\mathbf{e}_N\right) \thinspace+\mathbf{b}_i^l\textrm{.}
\end{split}
\end{equation}
To estimate the first-order partial derivatives of a vector-valued function, we apply the concept of Jacobian matrix, which is a matrix containing all the partial derivatives. For the vector-valued function $f:\mathbb{R}^N\to\mathbb{R}^M$, the Jacobian matrix is defined as $\left[\frac{\partial f}{\partial \bm{x}}\right]_{ij}= \frac{\partial}{\partial x_{j}}f(\bm{x})_i\in\mathbb{R}^{N\times M}$.
The partial derivative of C with respect to $\mathbf{b}_{i}^{l}$ is as follows:
\renewcommand{\arraystretch}{2}
\begin{equation}
\label{eq:derivative_b_fun}
\begin{split}
\frac{\partial C}{\partial \mathbf{b}_{i}^l}
&=
\begin{bmatrix}
\displaystyle
\frac{\partial C}{\partial b_{i1}^l} \textrm{,}~ 
\frac{\partial C}{\partial b_{i2}^l} \textrm{,}~\dots\textrm{,}~
\frac{\partial C}{\partial b_{iN}^l}
\end{bmatrix} \\
&=
\begin{bmatrix}
\displaystyle \frac{\partial C}{\partial z_{i1}^l} \frac{\partial z_{i1}^l}{\partial b_{i1}^l}+  
\displaystyle \frac{\partial C}{\partial z_{i2}^l} \frac{\partial z_{i2}^l}{\partial b_{i1}^l}+ 
\displaystyle \dots+
\displaystyle \frac{\partial C}{\partial z_{iN}^l} \frac{\partial z_{iN}^l}{\partial b_{i1}^l} \\ 
\displaystyle \vdots   \\
\displaystyle \frac{\partial C}{\partial z_{i1}^l} \frac{\partial z_{i1}^l}{\partial b_{iN}^l}+  
\displaystyle \frac{\partial C}{\partial z_{i2}^l} \frac{\partial z_{i2}^l}{\partial b_{iN}^l}+
\displaystyle \dots+
\displaystyle \frac{\partial C}{\partial z_{iN}^l} \frac{\partial z_{iN}^l}{\partial b_{iN}^l} \\
\end{bmatrix}^T \textrm{,}
\end{split}
\end{equation}
where $\frac{\partial C}{\partial \mathbf{b}_{i}^l}$ stands for the gradient vector of $\mathbf{b}_{i}^l$ belonging to neuron $i$ in layer $l$. The terms $\frac{\partial C}{\partial z_{in}^l}$ ($1 \leq n \leq N$)
are extracted into an $N$-dimensional local gradient vector $\frac{\partial C}{\partial \mathbf{z}_{i}^l}$ and we call it $\mathbf{d}_{i}^{l}$ for convenience:
\begin{equation}
\label{eq:delta_l}
\renewcommand{\arraystretch}{1}
\begin{split}
\frac{\partial C}{\partial \mathbf{z}_{i}^l}
&=\begin{bmatrix}
\displaystyle \frac{\partial C}{\partial z_{i1}^l}\textrm{,}~
\displaystyle \frac{\partial C}{\partial z_{i2}^l}\textrm{,}~
\dots \textrm{,}~
\displaystyle \frac{\partial C}{\partial z_{iN}^l}
\end{bmatrix}\\
&\triangleq\begin{bmatrix}
d_{i1}^l\textrm{,}~
d_{i2}^l\textrm{,}~
\dots\textrm{,}~
d_{iN}^l
\end{bmatrix}
=\mathbf{d}_{i}^{l} \textrm{.}
\end{split}
\end{equation}
The terms $\left\{\frac{\partial z_{i1}^{l}}{\partial b_{in}^{l}}\textrm{,}~\frac{\partial z_{i2}^{l}}{\partial b_{in}^{l}}\textrm{,}~\dots\textrm{,}~\frac{\partial z_{iN}^{l}}{\partial b_{in}^{l}}\right\}$ ($1 \leq n \leq N$) can be obtained by differentiating Eq.~\ref{eq:feedforward_extend_part2}:
\renewcommand{\arraystretch}{1.2}
\begin{equation}
\label{eq:partial_z_il_b_in}
\begin{split}
\frac{\partial \mathbf{z}_{i}^l}{\partial b_{in}^l}
&=
\begin{bmatrix}
\displaystyle
\frac{\partial z_{i1}^l}{\partial b_{in}^l} \textrm{,}~ 
\frac{\partial z_{i2}^l}{\partial b_{in}^l} \textrm{,}~\dots\textrm{,}~
\frac{\partial z_{iN}^l}{\partial b_{in}^l}
\end{bmatrix}^T \\
&=
\begin{bmatrix}
\displaystyle
\frac{\partial b_{i1}^l}{\partial b_{in}^l} \textrm{,}~ 
\frac{\partial b_{i2}^l}{\partial b_{in}^l} \textrm{,}~\dots\textrm{,}~
\frac{\partial b_{iN}^l}{\partial b_{in}^l}
\end{bmatrix}^T \\
&=
\underbrace{
\begin{bmatrix}
0 \textrm{,}~
\dots \textrm{,}~
1 \textrm{,}~
\dots \textrm{,}~
0 
\end{bmatrix}^T 
}_{\textrm{the }n \textrm{-th entry is }1}\textrm{.}
\end{split}
\end{equation}
By combining Eqs.~\ref{eq:delta_l} and \ref{eq:partial_z_il_b_in}, the derivative $\frac{\partial C}{\partial b_{in}^l}$ can be formulated as a dot product:
\renewcommand{\arraystretch}{1.0}
\begin{equation}
\label{eq:partial_C_bin_result}
\begin{split}
\frac{\partial C}{\partial b_{in}^l}
&=
\mathbf{d}_i^l\left[\frac{\partial \mathbf{z}_i^l}{\partial b_{in}^l}\right] =
\underbrace{
\begin{bmatrix}
d_{i1}^l\textrm{,}~
d_{i2}^l\textrm{,}~ 
\dots\textrm{,}~ 
d_{iN}^l
\end{bmatrix} 
}_{1 \times N} 
\underbrace{
\begin{bmatrix}
0 \\
\vdots \\
1 \\
\vdots \\ 
0 
\end{bmatrix}
}_{N \times 1} 
=
d_{in}^l \textrm{.}
\end{split}
\end{equation}
Then, each entry of $\frac{\partial C}{\partial \mathbf{b}_{i}^l}$ can be estimated by the same procedure from Eqs.~\ref{eq:partial_z_il_b_in} and \ref{eq:partial_C_bin_result}. After each entry is estimated, Eq.~\ref{eq:derivative_b_fun} can be derived as follows:
\renewcommand{\arraystretch}{2}
\begin{equation}
\label{eq:partial_C_over_partail_bi}
\begin{split}
\frac{\partial C}{\partial \mathbf{b}_{i}^l}
&=
\begin{bmatrix}
\displaystyle
\frac{\partial C}{\partial b_{i1}^l} \textrm{,}~ 
\dots\textrm{,} ~
\frac{\partial C}{\partial b_{iN}^l}
\end{bmatrix} =
\begin{bmatrix}
d_{i1}^{l} \textrm{,}~
\dots \textrm{,}~
d_{iN}^{l}
\end{bmatrix}
= \mathbf{d}_{i}^{l}  \textrm{.}
\end{split}
\end{equation}
As a result, it is evident that the derivative of $\mathbf{b}_{i}^l$ is completely equal to local gradient vector $\mathbf{d}_{i}^{l}$. 

Next, we derive the partial derivative of $C$ with respect to $\mathbf{w}_{ij}^l$, which is as follows: 
\renewcommand{\arraystretch}{2}
\begin{equation}
\label{eq:derivative_w_fun}
\begin{split}
\frac{\partial C}{\partial \mathbf{w}_{ij}^l}
&=
\begin{bmatrix}
\displaystyle
\frac{\partial C}{\partial w_{ij1}^l} \textrm{,}~ 
\frac{\partial C}{\partial w_{ij2}^l} \textrm{,}~\dots\textrm{,}~
\frac{\partial C}{\partial w_{ijN}^l}
\end{bmatrix} \\
&=
\begin{bmatrix}
\displaystyle \frac{\partial C}{\partial z_{i1}^l} \frac{\partial z_{i1}^l}{\partial w_{ij1}^l}+  
\displaystyle \frac{\partial C}{\partial z_{i2}^l} \frac{\partial z_{i2}^l}{\partial w_{ij1}^l}+ 
\displaystyle \dots+
\displaystyle \frac{\partial C}{\partial z_{iN}^l} \frac{\partial z_{iN}^l}{\partial w_{ij1}^l} \\ 
\displaystyle \frac{\partial C}{\partial z_{i1}^l} \frac{\partial z_{i1}^l}{\partial w_{ij2}^l}+  
\displaystyle \frac{\partial C}{\partial z_{i2}^l} \frac{\partial z_{i2}^l}{\partial w_{ij2}^l}+ 
\displaystyle \dots+
\displaystyle \frac{\partial C}{\partial z_{iN}^l} \frac{\partial z_{iN}^l}{\partial w_{ij2}^l} \\
\displaystyle \vdots   \\
\displaystyle \frac{\partial C}{\partial z_{i1}^l} \frac{\partial z_{i1}^l}{\partial w_{ijN}^l}+  
\displaystyle \frac{\partial C}{\partial z_{i2}^l} \frac{\partial z_{i2}^l}{\partial w_{ijN}^l}+
\displaystyle \dots+
\displaystyle \frac{\partial C}{\partial z_{iN}^l} \frac{\partial z_{iN}^l}{\partial w_{ijN}^l} \\
\end{bmatrix}^T 
\end{split}
\end{equation}
where $\frac{\partial C}{\partial \mathbf{w}_{ij}^l}$ stands for the gradient vector of $\mathbf{w}_{ij}^l$ connecting neuron $j$ to neuron $i$ in layer $l$ and it consists of $N$ elements. 
From Eq.~\ref{eq:delta_l}, here the terms $\left\{\frac{\partial C}{\partial z_{i1}^{l}}\textrm{,}~\frac{\partial C}{\partial z_{i2}^{l}}\textrm{,}~\dots\textrm{,}~\frac{\partial C}{\partial z_{iN}^{l}}\right\}$
are extracted as the local gradient vector $\mathbf{d}_{i}^{l}$.
Likewise, we can obtain the terms 
$\left\{\frac{\partial z_{i1}^l}{\partial w_{ijn}^{l}}\textrm{,}~\frac{\partial z_{i2}^l}{\partial w_{ijn}^{l}}\textrm{,}~\dots\textrm{,}~\frac{\partial z_{iN}^l}{\partial w_{ijn}^{l}}\right\}$ ($1 \leq n \leq N$)
by differentiating Eq.~\ref{eq:feedforward_extend_part2}:
\renewcommand{\arraystretch}{1.5}
\begin{equation}
\label{eq:partial_z_il_w_ijn}
\begin{split}
&\frac{\partial \mathbf{z}_{i}^l}{\partial w_{ijn}^l}
=
\begin{bmatrix}
\displaystyle
\frac{\partial z_{i1}^l}{\partial w_{ijn}^l} \textrm{,}~ 
\frac{\partial z_{i2}^l}{\partial w_{ijn}^l} \textrm{,}~\dots\textrm{,}~
\frac{\partial z_{iN}^l}{\partial w_{ijn}^l}
\end{bmatrix}^T \\
&=
a_{j1}^{l-1}\mathbf{e}_n\bullet\mathbf{e}_1+
a_{j2}^{l-1}\mathbf{e}_n\bullet\mathbf{e}_2+\dots+
a_{jN}^{l-1}\mathbf{e}_n\bullet\mathbf{e}_N\\
&=
\begin{bmatrix}
\mathbf{e}_n\bullet\mathbf{e}_1 \textrm{,}~
\mathbf{e}_n\bullet\mathbf{e}_2 \textrm{,} \dots \textrm{,}~
\mathbf{e}_n\bullet\mathbf{e}_N 
\end{bmatrix}
\begin{bmatrix}
a_{j1}^{l-1},
\dots 
a_{jN}^{l-1}
\end{bmatrix}^T \\
&=
\begin{bmatrix}
\mathbf{e}_n\bullet\mathbf{e}_1 \textrm{,}~
\mathbf{e}_n\bullet\mathbf{e}_2 \textrm{,} \dots \textrm{,}~
\mathbf{e}_n\bullet\mathbf{e}_N 
\end{bmatrix}
\mathbf{a}_{j}^{l-1} \textrm{,}
\end{split}
\end{equation}
which is formulated as a matrix-vector multiplication. Each column in the matrix is the bilinear product of two standard bases. Different bilinear products contribute to different results. Then, the derivative $\frac{\partial C}{\partial w_{ijn}^l}$ can be formulated as a dot product by combining Eqs.~\ref{eq:delta_l} and \ref{eq:partial_z_il_w_ijn}:
\renewcommand{\arraystretch}{1.0}
\begin{equation}
\label{eq:partial_C_over_wijn_result}
\begin{split}
\frac{\partial C}{\partial w_{ijn}^l}
&=
\mathbf{d}_i^l\left[\frac{\partial \mathbf{z}_i^l}{\partial w_{ijn}^l}\right]\\
&=
\underbrace{
\begin{bmatrix}
d_{i1}^l\textrm{,}~
d_{i2}^l\textrm{,}~ 
\dots\textrm{,}~ 
d_{iN}^l
\end{bmatrix} 
}_{1 \times N}
\\
&\quad\quad\quad\quad\quad
\left[
\underbrace{
\begin{bmatrix}
\mathbf{e}_n\bullet\mathbf{e}_1 \textrm{,}~
\mathbf{e}_n\bullet\mathbf{e}_2 \textrm{,}~
\dots \textrm{,}~
\mathbf{e}_n\bullet\mathbf{e}_N 
\end{bmatrix}
}_{N \times N}
\underbrace{
\mathbf{a}_{j}^{l-1} 
}_{N \times 1} 
\right]\\
&=
\mathbf{d}_i^l\left[\sum_{h=1}^{N}a_{jh}^{l-1}\mathbf{e}_n\bullet\mathbf{e}_h\right]\textrm{.}
\end{split}
\end{equation}
Each entry of $\frac{\partial C}{\partial \mathbf{w}_{ij}^l}$ can be estimated by the same procedure from Eqs.~\ref{eq:partial_z_il_w_ijn} and \ref{eq:partial_C_over_wijn_result}. After each entry is estimated, Eq.~\ref{eq:derivative_w_fun} can be rewritten as:
\renewcommand{\arraystretch}{2}
\begin{equation}
\label{eq:partial_C_over_partail_wij}
\begin{split}
\frac{\partial C}{\partial \mathbf{w}_{ij}^l}
&=
\begin{bmatrix}
\displaystyle
\frac{\partial C}{\partial w_{ij1}^l} \textrm{,}~ 
\frac{\partial C}{\partial w_{ij2}^l} \textrm{,}~\dots\textrm{,} ~
\frac{\partial C}{\partial w_{ijN}^l}
\end{bmatrix} \\
&=
\begin{bmatrix}
\displaystyle
\mathbf{d}_i^l\left[\sum_{h=1}^{N}a_{jh}^{l-1}\mathbf{e}_1\bullet\mathbf{e}_h\right] 
,\dots, 
\mathbf{d}_i^l\left[\sum_{h=1}^{N}a_{jh}^{l-1}\mathbf{e}_N\bullet\mathbf{e}_h\right]
\end{bmatrix}\\
&=
\mathbf{d}_{i}^{l}
\left[
\mathbf{e}_1\bullet\left[\sum_{h=1}^{N}a_{jh}^{l-1}\mathbf{e}_h\right]\textrm{,}~
\right.
\dots\textrm{,}
\left.\mathbf{e}_N\bullet\left[\sum_{h=1}^{N}a_{jh}^{l-1}\mathbf{e}_h\right]\right]\textrm{,}
\end{split}
\end{equation}
and referring to Eq.~\ref{eq:bilinear_product_first_term}, we can rewrite Eq.~\ref{eq:partial_C_over_partail_wij} into:
\begin{equation}
\label{eq:partial_C_over_partial_wij_result}
\begin{split}
\frac{\partial C}{\partial \mathbf{w}_{ij}^l}
&=
\mathbf{d}_{i}^{l}
\left[ 
\mathbf{e}_{1}\bullet\mathbf{a}_{j}^{l-1}\textrm{,}~
\mathbf{e}_{2}\bullet\mathbf{a}_{j}^{l-1}\textrm{,}~
\dots\textrm{,}~
\mathbf{e}_{N}\bullet\mathbf{a}_{j}^{l-1}
\right] \\
&=
\mathbf{d}_{i}^{l}\left[\mathbf{a}_{j}^{l-1}\right]_{\bullet}^{\dagger}\textrm{,}
\end{split}
\end{equation}
which is a vector-matrix multiplication with the transmuted representation.

After the derivative $\frac{\partial C}{\partial \mathbf{w}_{ij}^l}$ is calculated, we derive the local gradient vector $\mathbf{d}_{i}^l$ from layer $l+1$ by: 
\renewcommand{\arraystretch}{2.0}
\begin{equation}
\label{eq:delta_propagation}
\begin{split}
\mathbf{d}_{i}^{l}
=
\frac{\partial C}{\partial \mathbf{z}_{i}^l} 
&=
\sum_{k=1}^{K} \frac{\partial C}{\partial \mathbf{z}_{k}^{l+1}}
\frac{\partial \mathbf{z}_{k}^{l+1}}{\partial \mathbf{a}_{i}^{l}}
\frac{\partial \mathbf{a}_{i}^{l}}{\partial \mathbf{z}_{i}^l} \\
&=
\sum_{k=1}^{K} \frac{\partial C}{\partial \mathbf{z}_{k}^{l+1}} 
\frac{\partial \mathbf{z}_{k}^{l+1}}{\partial \mathbf{a}_{i}^{l}}
\dot{\phi}({\mathbf{z}_{i}^{l}}) \textrm{,}
\end{split}
\end{equation}
which is a vector-matrix-matrix multiplication, where $\mathbf{z}_{k}^{l+1}$ stands for an $N$-dimensional input vector of neuron $k$ in layer $l+1$ and $\mathbf{a}_{i}^{l}$ stands for an $N$-dimensional output vector of neuron $i$ in layer $l$. The vector $\frac{\partial C}{\partial \mathbf{z}_{k}^{l+1}}$ is regarded as an $N$-dimensional local gradient vector, defined as $\mathbf{d}_{k}^{l+1}$ in layer $l+1$. Referring to Eqs. \ref{eq:derivative_b_fun} to \ref{eq:partial_C_over_partail_bi}, we can see that $\mathbf{d}_{k}^{l+1}$ can be regarded as the derivative of $\mathbf{b}_{k}^{l+1}$. 
The matrix $\dot{\phi}({\mathbf{z}_{i}^{l}})$ represents the derivative of the activation function, which is an $N \times N$ diagonal matrix:
\renewcommand{\arraystretch}{1.5}
\begin{equation}
\begin{split}
\dot{\phi}({\mathbf{z}_{i}^{l}})
&=
\frac{\partial \mathbf{a}_{i}^{l}}{\partial \mathbf{z}_{i}^l} \\
&=
\begin{bmatrix}
\displaystyle \frac{\partial a_{i1}^{l}}{\partial z_{i1}^{l}}  & 0          & \dots	     & 0            \\ 
0          & \displaystyle \frac{\partial a_{i2}^{l}}{\partial z_{i2}^{l}}  & \dots        & 0          \\
\vdots     & \vdots     & \ddots       & \vdots                                                         \\
0          & 0          & \dots        & \displaystyle \frac{\partial a_{iN}^{l}}{\partial z_{iN}^{l}}  \\
\end{bmatrix} \\
&=
\begin{bmatrix}
\dot{\phi}({z_{i1}^{l}})  & 0                         & \dots	           & 0                         \\ 
0                         & \dot{\phi}({z_{i2}^{l}})  & \dots              & 0                         \\
\vdots                    & \vdots                    & \ddots             & \vdots                    \\
0                    	  & 0                    	  & \dots              & \dot{\phi}({z_{iN}^{l}})  \\
\end{bmatrix} \textrm{,}
\end{split}
\end{equation} 
where $\dot{\phi}$ can be an arbitrary differentiable activation function.
Next, the matrix $\frac{\partial \mathbf{z}_{k}^{l+1}}{\partial \mathbf{a}_{i}^{l}}$ is also a $N\times N$ square matrix:
\renewcommand{\arraystretch}{2}
\begin{equation}
\label{eq:partial_zk_over_partial_ai}
\begin{split}
\frac{\partial \mathbf{z}_{k}^{l+1}}{\partial \mathbf{a}_{i}^{l}} 
&=
\begin{bmatrix}
& \displaystyle \frac{\partial z_{k1}^{l+1}}{\partial a_{i1}^{l}}   
& \displaystyle \frac{\partial z_{k1}^{l+1}}{\partial a_{i2}^{l}} 
& \displaystyle \dots
& \displaystyle \frac{\partial z_{k1}^{l+1}}{\partial a_{iN}^{l}}  \\ 
& \displaystyle \frac{\partial z_{k2}^{l+1}}{\partial a_{i1}^{l}} 
& \displaystyle \frac{\partial z_{k2}^{l+1}}{\partial a_{i2}^{l}}  
& \displaystyle \dots
& \displaystyle \frac{\partial z_{k2}^{l+1}}{\partial a_{iN}^{l}}  \\
& \vdots   
& \vdots 
& \ddots
& \vdots  \\
& \displaystyle \frac{\partial z_{kN}^{l+1}}{\partial a_{i1}^{l}}  
& \displaystyle \frac{\partial z_{kN}^{l+1}}{\partial a_{i2}^{l}} 
& \displaystyle \dots
& \displaystyle \frac{\partial z_{kN}^{l+1}}{\partial a_{iN}^{l}}  \\
\end{bmatrix} \textrm{.}
\end{split}
\end{equation}
We calculate the above derivatives columnwise. The calculations are based on the feedforward process of $\mathbf{z}_k^{l+1}$:
\renewcommand{\arraystretch}{1.2}
\begin{equation}
\begin{split}
\label{eq:feedforward_extend_l+1}
\mathbf{z}_k^{l+1}
&=\sum_{i=1}^{I}\mathbf{w}_{ki}^{l+1}\bullet\mathbf{a}_{i}^{l}+\mathbf{b}_{k}^{l+1}\\
&=\sum_{i=1}^{I}\mathbf{w}_{ki}^{l+1}\bullet\left[a_{i1}^{l}\mathbf{e}_{1}+a_{i2}^{l}\mathbf{e}_{2}+\dots +a_{iN}^{l}\mathbf{e}_{N}\right]+\mathbf{b}_{k}^{l+1} \textrm{.}\\
\end{split}
\end{equation}
Then, the derivative of the $n$-th column of Eq.~\ref{eq:partial_zk_over_partial_ai} is derived by differentiating Eq.~\ref{eq:feedforward_extend_l+1}:
\renewcommand{\arraystretch}{1.5}
\begin{equation}
\begin{split}
\frac{\partial \mathbf{z}_k^{l+1}}{\partial a_{in}^l}
&=
\begin{bmatrix}
\displaystyle
\frac{\partial z_{k1}^{l+1}}{\partial a_{in}^l} \textrm{,} 
\frac{\partial z_{k2}^{l+1}}{\partial a_{in}^l} \textrm{,}\dots\textrm{,} 
\frac{\partial z_{kN}^{l+1}}{\partial a_{in}^l}
\end{bmatrix}^T \\
&=\mathbf{w}_{ki}^{l+1}\bullet\mathbf{e}_n \\
&=
\left(
w_{ki1}^{l+1}\mathbf{e}_1+w_{ki2}^{l+1}\mathbf{e}_2+\dots+w_{kiN}^{l+1}\mathbf{e}_N
\right)\bullet\mathbf{e}_n\\
&=
\sum_{h=1}^{N}w_{kih}^{l+1}\mathbf{e}_h\bullet\mathbf{e}_n \\
&=
\underbrace{
\left[
\mathbf{e}_1\bullet\mathbf{e}_n\textrm{,}~
\mathbf{e}_2\bullet\mathbf{e}_n\textrm{,}~
\dots~\textrm{,}
\mathbf{e}_N\bullet\mathbf{e}_n
\right]}_{N \times N}
\underbrace{
\mathbf{w}_{ki}^{l+1} 
}_{N \times 1} \textrm{,}
\end{split}
\end{equation}
resulting in a column vector. Since each column of the matrix $\frac{\partial \mathbf{z}_{k}^{l+1}}{\partial \mathbf{a}_{i}^l}$ can be estimated by the same procedure mentioned above, the derivatives of the matrix can be derived as:
\begin{equation}
\label{eq:partial_zk_over_partial_ai_result}
\begin{split}
\frac{\partial \mathbf{z}_{k}^{l+1}}{\partial \mathbf{a}_{i}^l}
&=
\left[
\left[\sum_{h=1}^{N}w_{kih}^{l+1}\mathbf{e}_h\bullet\mathbf{e}_1\right]\textrm{,}~
\left[\sum_{h=1}^{N}w_{kih}^{l+1}\mathbf{e}_h\bullet\mathbf{e}_2\right]\textrm{,} 
\right.\\
&\quad\quad\quad\quad\quad\quad\quad\quad\thinspace
\dots
\textrm{,}
\left.
\left[\sum_{h=1}^{N}w_{kih}^{l+1}\mathbf{e}_h\bullet\mathbf{e}_N\right]\right] \\
&=
\left[
\mathbf{w}_{ki}^{l+1}\bullet\mathbf{e}_1 \textrm{,}~
\mathbf{w}_{ki}^{l+1}\bullet\mathbf{e}_2 \textrm{,}~
\dots \textrm{,}~
\mathbf{w}_{ki}^{l+1}\bullet\mathbf{e}_N
\right] \\
&=
\left[\mathbf{w}_{ki}^{l+1}\right]_{\bullet}\textrm{,}
\end{split}
\end{equation}
in which the matrix is made up of $N$ vectors. After estimating the matrix $\frac{\partial \mathbf{z}_{k}^{l+1}}{\partial \mathbf{a}_{i}^l}$, Eq.~\ref{eq:delta_propagation} can be rewritten as follows:
\begin{equation}
\label{eq:delta_propagation_new}
\begin{split}
\mathbf{d}_{i}^{l}
&=
\sum_{k=1}^{K}\mathbf{d}_{k}^{l+1}\left[\mathbf{w}_{ki}^{l+1}\right]_{\bullet}
\dot{\phi}({\mathbf{z}_{i}^{l}})\textrm{,}
\end{split}
\end{equation}
which becomes a vector-matrix-vector multiplication. It is evident that local gradient vector $\mathbf{d}_{i}^{l}$ in layer $l$ is backpropagated from $\mathbf{d}_{k}^{l+1}$ in layer $l+1$. From this we see that the local gradient can be inferred from the output layer. In the output layer $L$, the local gradient vector $\mathbf{d}_{g}^{L}$ is defined as:
\begin{equation}
\label{eq:delta_calculation_at_layer_L}
\begin{split}
\frac{\partial C}{\partial \mathbf{z}_{g}^{L}}
&=\begin{bmatrix}
\displaystyle
\frac{\partial C}{\partial z_{g1}^{L}} \textrm{,}~
\frac{\partial C}{\partial z_{g2}^{L}} \textrm{,}~
\dots\textrm{,}~
\frac{\partial C}{\partial z_{gN}^{L}} 
\end{bmatrix} \\
&=\begin{bmatrix}
\displaystyle
\frac{\partial C}{\partial y_{g1}^{L}} \frac{\partial y_{g1}^{L}}{\partial z_{g1}^{L}} \textrm{,}~
\frac{\partial C}{\partial y_{g2}^{L}} \frac{\partial y_{g2}^{L}}{\partial z_{g2}^{L}} \textrm{,}~
\dots\textrm{,}~
\frac{\partial C}{\partial y_{gN}^{L}} \frac{\partial y_{gN}^{L}}{\partial z_{gN}^{L}}
\end{bmatrix} \\
&=\begin{bmatrix}
\displaystyle
\frac{\partial C}{\partial y_{g1}^{L}} \dot{\phi}({z_{g1}^{L}}) \textrm{,}~
\frac{\partial C}{\partial y_{g2}^{L}} \dot{\phi}({z_{g2}^{L}}) \textrm{,}~
\dots\textrm{,}~
\frac{\partial C}{\partial y_{gN}^{L}} \dot{\phi}({z_{gN}^{L}})
\end{bmatrix} \textrm{,}
\end{split}
\end{equation}
where $\frac{\partial C}{\partial y_{gn}^{L}}$ ($1 \leq n \leq N$) is the derivative of the loss function used in the feedforward process, and $\displaystyle \dot{\phi}({z_{gn}^{L}})$ is the derivative of the activation function. The above process is summarized in Algorithm~\ref{alg:ABIPalg}.

\begin{algorithm}[t]
\caption{Arbitrary Bilinear Product Backpropagation}
\begin{algorithmic}[1]
\REQUIRE Training inputs $\left\{\mathbf{X}_{m}\right\}_{m=1}^{M}$ \\
		\hspace{0.42cm} Training targets $\left\{\mathbf{Y}_{m}\right\}_{m=1}^{M}$ \\
		\hspace{0.42cm} Bilinear product $\bullet$ \\
		\hspace{0.42cm} Activation function $\phi$
\ENSURE  Parameters $\mathbf{\Theta}=\left\{\mathbf{\mathcal{W}}^{l}, \mathbf{\mathcal{B}}^{l}\right\}_{l=1}^{L}$ 

\WHILE{not converged}
\FOR{\textbf{each} minibatch  $\in \left\{\mathbf{X}\right\}_{m=1}^{M}$}
    \STATE Compute $\left\{\mathbf{\mathcal{Z}}^{l},\mathbf{\mathcal{A}}^{l}\right\}_{l=1}^{L}$ with Eqs.~\ref{eq:feedforward}--\ref{eq:feedforeard_at_layer_L} (feedforward) 
  \STATE Compute $\left\{\mathbf{\mathcal{D}}^{l}\right\}^{L}_{l=1}$ with Eqs.~\ref{eq:delta_propagation_new} and \ref{eq:delta_calculation_at_layer_L} (backprop)
  \STATE Update $\left\{\mathbf{\mathcal{W}}^{l}\right\}^{L}_{l=1}$ with the gradients from Eq.~\ref{eq:partial_C_over_partial_wij_result}
  \STATE Update $\left\{\mathbf{\mathcal{B}}^{l}\right\}^{L}_{l=1}$ with the gradients from Eq.~\ref{eq:partial_C_over_partail_bi}
\ENDFOR 
\ENDWHILE 
\end{algorithmic}
\label{alg:ABIPalg}
\end{algorithm}

\subsection{Example: Circular Convolution as The Bilinear Product}
\label{subsec:circular_convolution_induction}
In the above, we generalized the vector NNs using arbitrary bilinear products.  This algorithm can be realized to train model parameters for scalar neurons or vector neurons from training data based on matrices or tensors. In this subsection, we derive $\left[\mathbf{a}_{j}^{l-1}\right]_{\bullet}^{\dagger}$ and $\left[\mathbf{w}_{ki}^{l+1}\right]_{\bullet}$ for circular convolution. The derivation is described between layer $l-1$ and layer $l$. 

Suppose $\mathbf{w}_{ij}^{l}\bullet\mathbf{a}_{ij}^{l-1}$ stands for the usual circular convolution, $\mathbf{w}_{ij}^{l}=\left[w_{ij1}^l\textrm{,}~w_{ij2}^l\textrm{,}\dots\textrm{,}~w_{ijN}^l\right]^T\in \mathbb{R}^N$ and $\mathbf{a}_{j}^{l-1}=\left[a_{j1}^{l-1}\textrm{,}~a_{j2}^{l-1}\textrm{,}\dots\textrm{,}~a_{jN}^{l-1}\right]^T\in \mathbb{R}^N$ are both $N$-dimensional vectors, and the matrix representation of $\mathbf{w}_{ij}^{l}\bullet\mathbf{a}_{ij}^{l-1}$ is:
\renewcommand{\arraystretch}{1.5}
\begin{equation}
\label{eq:circular_convolution}
\begin{split}
\left[\mathbf{w}_{ij}^{l}\right]_{\bullet}=
\begin{bmatrix}
w_{ij1}^l  & w_{ijN}^l      & w_{ij(N-1)}^l  & \dots     & w_{ij2}^l   \\ 
w_{ij2}^l  & w_{ij1}^l      & w_{ijN}^l      & \dots     & w_{ij3}^l   \\
w_{ij3}^l  & w_{ij2}^l      & w_{ij1}^l      & \dots     & w_{ij4}^l   \\
\vdots     & \vdots         & \vdots         & \ddots    & \vdots      \\
w_{ijN}^l  & w_{ij(N-1)}^l  & w_{ij(N-2)}^l  & \dots     & w_{ij1}^l   \\
\end{bmatrix} \textrm{,}
\end{split}
\end{equation}
where the weight vector is formulated as an $N \times N$ square matrix with $w_{ij1}^l$ on the main diagonal. The matrix $\left[\mathbf{a}_{j}^{l-1}\right]_{\bullet}^{\dagger}$ for Eq.~\ref{eq:partial_C_over_partial_wij_result} is extended as follows:
\begin{equation}
\begin{split}
\left[\mathbf{a}_{j}^{l-1}\right]_{\bullet}^{\dagger} =
\begin{bmatrix}
a_{j1}^{l-1}  & a_{jN}^{l-1}      & a_{j(N-1)}^{l-1}  & \dots     & a_{j2}^{l-1}  \\ 
a_{j2}^{l-1}  & a_{j1}^{l-1}      & a_{jN}^{l-1}      & \dots     & a_{j3}^{l-1}  \\
a_{j3}^{l-1}  & a_{j2}^{l-1}      & a_{j1}^{l-1}      & \dots     & a_{j4}^{l-1}  \\
\vdots        & \vdots            & \vdots            & \ddots    & \vdots        \\
a_{jN}^{l-1}  & a_{j(N-1)}^{l-1}  & a_{j(N-2)}^{l-1}  & \dots     & a_{j1}^{l-1}  \\
\end{bmatrix} \textrm{,}
\end{split}
\end{equation}
where the permutation of elements in the matrix $\left[\mathbf{a}_{j}^{l-1}\right]_{\bullet}^{\dagger}$ is identical to the matrix $\left[\mathbf{w}_{ij}^{l}\right]_{\bullet}$. The matrix $\left[\mathbf{w}_{ki}^{l+1}\right]_{\bullet}$ for circular convolution is extended as:
\begin{equation}
\begin{split}
\left[\mathbf{w}_{ki}^{l+1}\right]_{\bullet}=
\begin{bmatrix}
w_{ki1}^{l+1}  & w_{kiN}^{l+1}      & w_{ki(N-1)}^{l+1}  & \dots     & w_{ki2}^{l+1}   \\ 
w_{ki2}^{l+1}  & w_{ki1}^{l+1}      & w_{kiN}^{l+1}      & \dots     & w_{ki3}^{l+1}   \\
w_{ki3}^{l+1}  & w_{ki2}^{l+1}      & w_{ki1}^{l+1}      & \dots     & w_{ki4}^{l+1}   \\
\vdots         & \vdots             & \vdots             & \ddots    & \vdots          \\
w_{kiN}^{l+1}  & w_{ki(N-1)}^{l+1}  & w_{ki(N-2)}^{l+1}  & \dots     & w_{ki1}^{l+1}   \\
\end{bmatrix} \textrm{.}
\end{split}
\end{equation}
Note that when $N$ equals two, this architecture becomes the hyperbolic NN \cite{buchholz00ijcnn}. When $N$ is larger than two, this product is also known as polar complex multiplication \cite{olariu02elsevier,chan16tsp}. Detail investigations of more bilinear products are described in Appendix~\ref{appendix:bilinear_product_induction}.

\section{Experiments}
\label{sec:experiments}
To validate the effectiveness of ABIPNN, we consider two regression problems that require learning a nonlinear mapping between tensor inputs and tensor outputs. The first one is \emph{multispectral image denoising}, which aims to recover the original multispectral image from an noisy input. 
The second one is \emph{blind singing voice separation}, which aims to separate the singing voice and the accompaniment from a monaural audio mixture.
The dimensionality $N$ of the data we consider for the two problems are $10$ and $3$, respectively. 

We intend to empirically compare the performance of the conventional DNNs with ABIPNN. For both DNNs and ABIPNN, we use sigmoid as the activation function, mean-square error (MSE) as the objective function, and Adam~\cite{kingma14arxiv} for gradient optimizations, with a learning rate of $5\times10^{-4}$. 
For ABIPNN, we employ circular convolution as our bilinear product. 
The experiments are performed in MATLAB on a personal computer equipped with an NVIDIA GeForce GTX $1080$ Ti GPU and a memory of $64$ GB RAM. For reproducible research, we will make the code of the experiments publicly available at a project website.

\subsection{Experiment on Multispectral Image Denoising}
\label{subsec:multispectral_image_denoising}
Multispectral (a.k.a.~hyperspectral) imaging systems are usually employed to solve broadband color problems. A multispectral image is composed of a collection of monospectral (or monochrome) images, each of which is captured with a specific wavelength. These monospectral images can be considered as in different \emph{bands} of the multispectral image. As different spectral bands may exhibit some associations among each other, we can leverage such associations to enhance the accuracy of image processing applications. 

In multispectral image denoising, we are given a noisy version of a multispectral image  with $N$ bands, and are asked to recover the clean version (also $N$ bands). In a recent work presented by Zhang~\emph{et al.}~\cite{zhang16ijcai}, different supervised multidimensional dictionary learning methods were evaluated on the Columbia multispectral image database~\cite{yasuma10itip} for the denoising task. By following their settings, we can compare the performance of our models with these prior arts.
These methods include
K-TSVD, K-SVD~\cite{aharon06sp}, $3$D K-SVD~\cite{elad06ip},  
LRTA~\cite{renard08grs}, DNMDL~\cite{peng14cvpr} and PARAFAC~\cite{liu12grs}. 

The database~\cite{yasuma10itip} contains $32$ real-world scenes and each scene contains $31$ monochrome images of size $512\times512$, captured by varying the wavelength of a camera from $400$ nm to $700$ nm with a step size of $10$ nm. Following Zhang~\emph{et al.}~\cite{zhang16ijcai}, we consider images in the ``chart and stuffed toy'' scene, resize each image to $205\times205$ and take images of the last $10$ bands
(i.e., starting from $600$ nm), making $N=10$.
Moreover, we divide each image into $8\times8\times10$ overlapping tensor patches with a hop size of one. We randomly take $10\,000$ tensor patches as the training data and the rest for testing. 
From the training data, $1000$ tensor patches are held out as the validation data.  
The maximal number of training epochs is set to $3000$. We also stop the training process if the validation MSE does not decrease for $100$ epochs.

\begin{figure}[t]
\begin{center}
\includegraphics[trim=3cm 5cm 3cm 4cm, clip, scale=0.35]{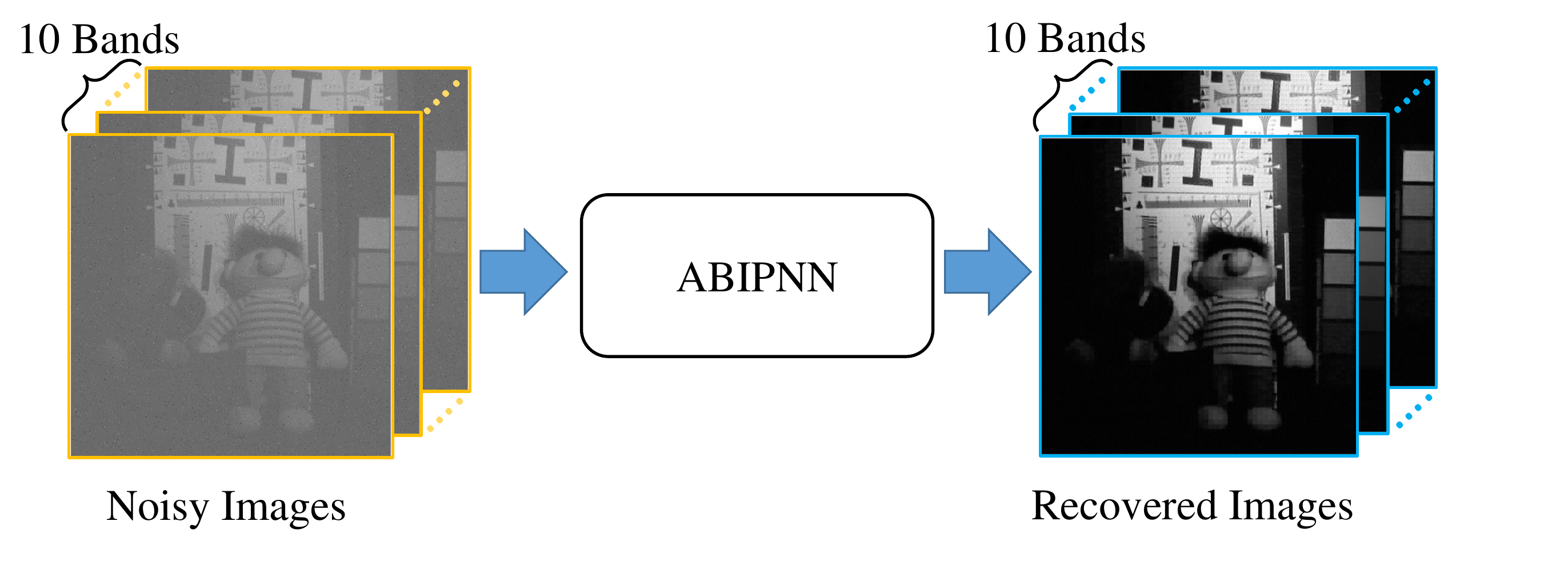}
\caption{Illustration of multispectral image denoising using ABIPNN. Noisy images are contaminated by Gaussian noise. We take the last $10$ bands for the experiments.}
\label{fig:multispectral_image_denosing_arch}
\end{center}
\end{figure}

For the noise model, we randomly select a certain number of pixels from each band of an image and add to the pixels Gaussian noise with specific \emph{sigma value}.
We refer to the ratio of pixels corrupted per band as the \emph{sparisty} level of the noise.
In our experiments, we vary the sigma value from $100$ to $200$, and the sparsity level from $5\%$ to $15\%$, to simulate different degrees of corruption. 
The goal is to recover the corrupted images, as illustrated in 
Fig. \ref{fig:multispectral_image_denosing_arch}.
As for the objective function in model training, we compute the MSE between the recovered and the original versions of the patches in the training set, across all the $10$ bands. 
As Zhang~\emph{et al.}~\cite{zhang16ijcai},  we measure the performance of denoising in terms of the peak signal-to-noise ratio (PSNR). We calculate the PSNR for each test patch and report the average result.

In our implementation, the ABIPNN consists of $3$ hidden layers and $512$ neurons in each layer (i.e., $J=I=K=512$ in Fig.~\ref{fig:vnn_arch}). Both the input and output layers have $8\times8=64$ dimensions (i.e., $R=G=64$ in Fig.~\ref{fig:vnn_arch}). We denote such a topology as $64$-$512$-$512$-$512$-$64$ hereafter. In ABIPNN, each neuron represents a vector with size $N=10$. Hence, for each patch the $10$ bands are processed at the same time.
To compare the performance of ABIPNN with conventional DNNs using similar number of parameters, we consider the following two variants of DNNs as the baselines. 
The first variant, \emph{DNN-concat}, simply concatenates the $10$ bands as a single vector to process them jointly, making $R=G=640$. There are $3$ hidden layers and $1,450$ neurons in each layer, so the topology is $640$-$1450$-$1450$-$1450$-$640$, where each neuron represents a scalar.
The second variant, \emph{DNN-parallel}, processes the $10$ bands separately using $10$ DNNs, each having a $64$-$512$-$512$-$512$-$64$ topology. 
In other words, each DNN is trained for denoising a specific band.

\begin{figure}[t]
\begin{center}
``Chart and Stuffed Toy'' Image
\includegraphics[trim=2cm 4cm 1.5cm 4cm, clip, scale=0.75]{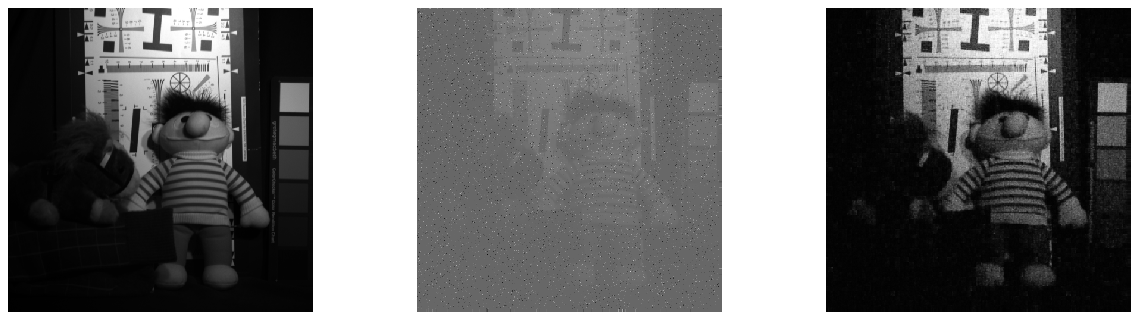}
``Face'' Image
\includegraphics[trim=2cm 4cm 1.5cm 4cm, clip, scale=0.75]{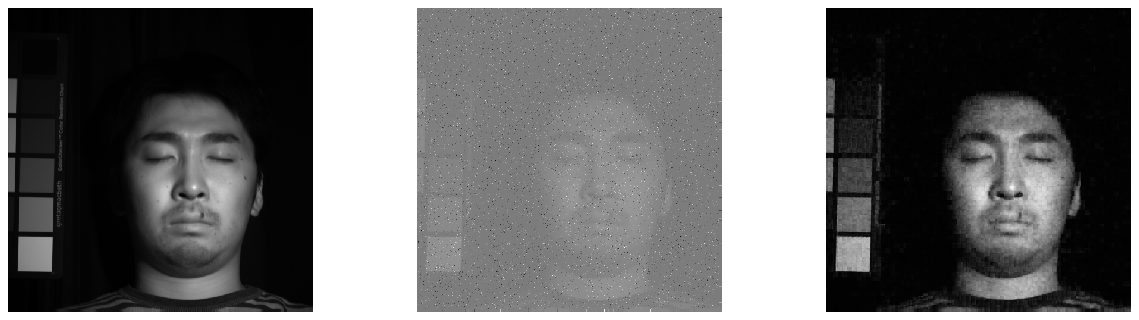}
``Paints'' Image
\includegraphics[trim=2cm 4cm 1.5cm 4cm, clip, scale=0.75]{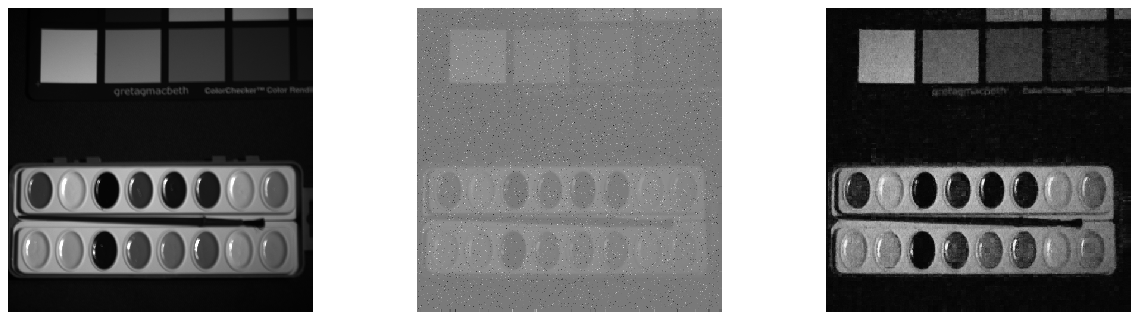}
``Fake and Real Peppers'' Image
\includegraphics[trim=2cm 4cm 1.5cm 4cm, clip, scale=0.75]{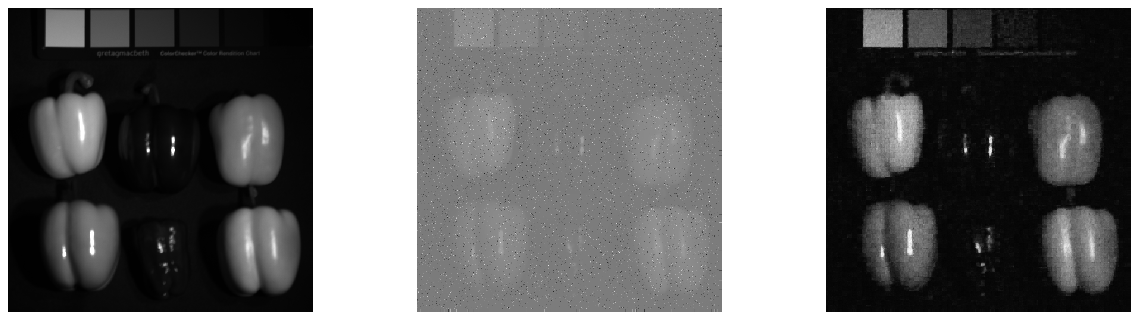}

(a) Original ~~~~~~~~~~~~ (b) Noisy ~~~~~~~~~~~~ (c) Recovered 
\caption{Denoised images at the $700$ nm band using the proposed ABIPNN method. 
The sparsity of the noisy pixels is $10$\% and the sigma value of the additive Gaussian noise is set to $200$.}
\label{fig:multispectral_image_denosing_examples}
\end{center}
\end{figure}

Table~\ref{tab:multispectral_image_denosing_table} shows the experimental results. In the upper part of the table, we cite the PSNR values reported in \cite{zhang16ijcai}, and in the lower part we report the results of our own implementation. 
Because we follow their experimental settings, the PSNR values for the noisy images (i.e., before denoising) reported in \cite{zhang16ijcai} are close to what we observe in our implementation.
Besides, Table~\ref{tab:multispectral_image_denosing_table} also shows that
ABIPNN outperforms all the other methods, including the two DNN baselines, by a large margin across different values of sigma and sparsity. 
The PSNRs are improved from 12.10--20.92 dB to 29.55--33.92 dB.
This result suggests the effectiveness of a neural network for this task.
Figure~\ref{fig:multispectral_image_denosing_examples} demonstrates the original, noisy, and denoised versions of three images by using ABPINN. 

\begin{table}[t]
\captionsetup{justification=centering,margin=0.5cm}
\caption{\textsc{PSNR (in dB) obtained by different methods for multispectral image denoising, under different sparsity and sigma values}}
\label{tab:multispectral_image_denosing_table}
\begin{tabular}{|l|c|c|c|c|c|}
\hline
Sparsity        & $5\%$          & $10\%$          & $15\%$          & $10\%$          & $10\%$  \\ 
Sigma           & $100$          & $100$           & $100$           & $150$           & $200$   \\
\hline
\multicolumn{6}{|l|}{\textbf{Referenced from \cite{zhang16ijcai}}} \\
\hline
Noisy Image     & $20.96$        & $18.18$         & $16.35$         & $14.75$         & $12.10$ \\ 
K-SVD~\cite{aharon06sp}          & $22.73$        & $22.60$         & $22.49$         & $22.38$         & $22.20$ \\
$3$DK-SVD~\cite{elad06ip}      & $22.61$        & $22.53$         & $22.47$         & $22.41$         & $22.20$ \\
LRTA~\cite{renard08grs}           & $23.54$        & $26.84$         & $26.65$         & $23.90$         & $22.03$ \\
DNMDL~\cite{peng14cvpr}          & $24.07$        & $23.73$         & $25.16$         & $17.89$         & $16.83$ \\
PARAFAC~\cite{liu12grs}        & $27.07$        & $26.86$         & $26.72$         & $26.13$         & $25.24$ \\
K-TSVD~\cite{zhang16ijcai}         & $27.19$        & $26.98$         & $26.79$         & $26.18$         & $25.44$ \\
\hline
\multicolumn{6}{|l|}{\textbf{Our Experiments}} \\
\hline
Noisy Image     & $20.92$        & $18.16$         & $16.35$         & $14.64$         & $12.10$ \\
DNN-concat & $25.06$        & $24.80$         & $24.93$         & $24.59$         & $24.03$ \\
DNN-parallel & $30.18$        & $28.88$         & $28.06$         & $27.17$         & $25.88$ \\ 
ABIPNN   & $ {33.92}$   & $ {32.47}$    & $ {31.74}$    & $ {31.01}$    & $ {29.55}$ \\
\hline
\end{tabular}
\end{table}

\begin{figure}[t]
\begin{center}
\includegraphics[trim=0.1cm 0.05cm 0.5cm 0.1cm, clip, scale=0.63]{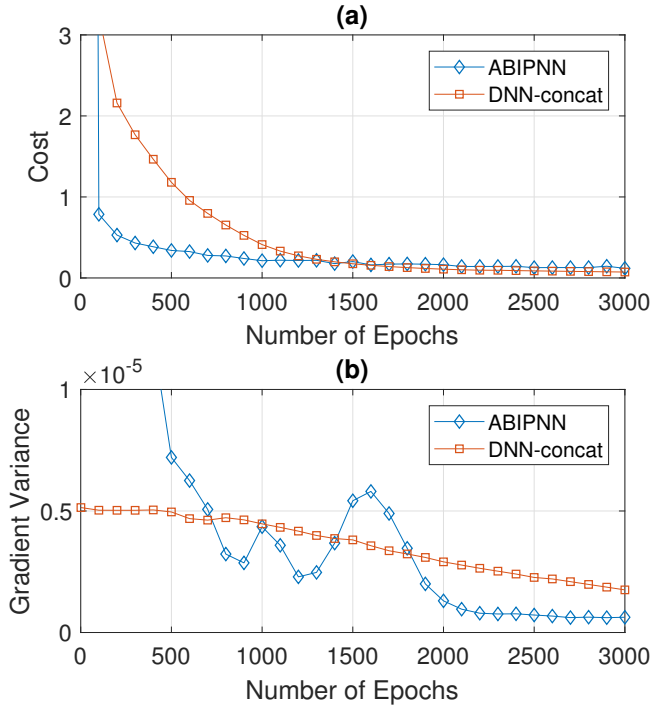}
\caption{(a) Mean square error with Adam and (b) gradient variance with SGD in the training procedure of ABIPNN and DNN-concat, for the case when the sparsity of the noisy pixels is $5$\% and the sigma value is set to $100$.}
\label{fig:error_cost_gradient_variance_trend}
\end{center}
\end{figure}

Among the three NN-based methods, DNN-concat performs the worst. Its performance is even inferior to that of PARAFAC~\cite{liu12grs} and K-TSVD~\cite{zhang16ijcai}, two non-deep learning based methods. This suggests that concatenating inputs from different bands does not make it easy for an NN to learn the association between bands. 
In contrast, using multidimensional vector neurons, ABIPNN learns the relations between bands by computing the circular convolution of two vectors: one coming from a hidden node and the other coming from a weight tensor. The vector coming from a weight tensor can be regarded as a linear kernel that captures interactions across bands, which may contribute to 
enhanced results in denoising.



Figure~\ref{fig:error_cost_gradient_variance_trend}(a) displays the changes in MSE values as a function of the training epochs in the training procedure for ABIPNN and DNN-concat. We find that the MSE values converge to a certain value for both methods, but ABIPNN converges much faster. We conjecture that this is because the error propagation in ABIPNN is not only optimized for each dimension (i.e., band) but also between different dimensions.
It is also known that gradient variance reduction helps achieve better convergence in stochastic gradient descent (SGD)~\cite{johnson13nips,reddi15nips}.
Fig.~\ref{fig:error_cost_gradient_variance_trend}(b) shows the changes in gradient variance during SGD training. Here ABIPNN provides lower variance at convergence (after $2000$ epochs).

\subsection{Experiment on Blind Singing Voice Separation} 
\label{subsec:singing_voice_separation}

Separating the singing voice and the accompaniment from a monaural audio mixture is challenging, because there are more unknowns than equations. 
Unsupervised methods such as robust principal component analysis \cite{huang12icassp, yang13ismir, chan15icassp} assume no labeled data are available and rely on  assumptions on the characteristics of the sources for separation. For example, the spectrogram of the accompaniment part is assumed to have lower rank than that of the vocal part.
If we are given the original sources of some audio mixtures, we can take these clean sources as the supervisory signal and employ supervised  methods such as non-negative matrix factorization (NMF) \cite{lee99nature}  for source separation.
Naturally, supervised methods usually outperform unsupervised methods, as the former can learn from the pairs of mixtures and sources.
Recently, NN-based methods have been introduced to this task \cite{huang14ismir,hershey16icassp}, showing better result than non-NN based methods such as NMF.
This can be done by taking the spectrogram of an audio mixture as the input and requiring the network to reproduce the spectrograms of the corresponding two sources at the output.  NN works better because they can learn a nonlinear mapping between the input and the outputs.

A spectrogram is a 2-D time-frequency representation.  It is computed by the short time Fourier transform (STFT), which divides a given time signal into short segments of equal length and then computes the Fourier transform separately on each short segment. We call each short segment a `frame,' and the result of Fourier transform per frame as  a `spectrum.'  The spectrogram considers only the magnitude part of STFT.

\begin{figure}[t]
\begin{center}
\includegraphics[trim=4.5cm 0.5cm 5.8cm 1.4cm, clip, scale=0.40]{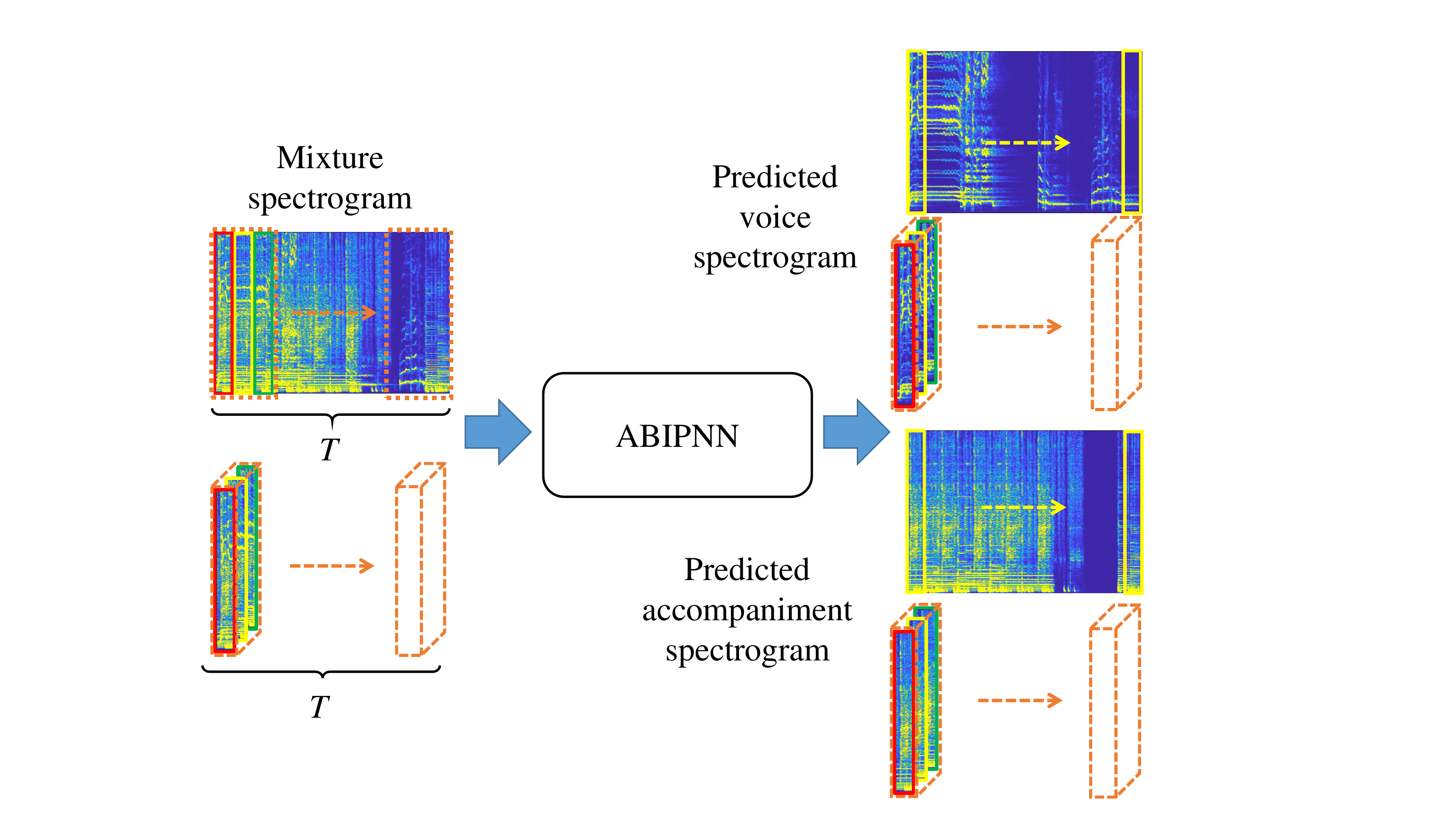}
\caption{Illustration (best seen in color) of singing voice separation using ABIPNN, where $T$ stands for the number of frames. We use red to indicate the previous frames, yellow for the current frames, and green for the subsequent frames. After training, we extract the second dimension (yellow) for soft-time frequency masking.}
\label{fig:singing_voice_separation_arch}
\end{center}
\end{figure}

\begin{table*}[t]
\centering
\captionsetup{justification=centering,margin=0.5cm}
\caption{\textsc{Comparison of SDR, SIR and SAR values by using different models under different topologies number of parameters on DSD100 data set}}
\label{tab:dsd100_result_with_nn}
\begin{tabular}{|l|c|l|c|cc|cc|cc|}
\hline
\multirow{2}{*}{Model} &\multirow{2}{*}{Window} &\multirow{2}{*}{Topology} &Number of  &\multicolumn{2}{c}{SDR} &\multicolumn{2}{c}{SIR} &\multicolumn{2}{c|}{SAR} \\
\cline{5-10}
                   &          &                          &   parameters & Vocal   &Accomp.      & Vocal   &Accomp.      & Vocal   &Accomp.   \\
\hline
 DNN-simple            & $1$      & $513$-$513$-$513$-$513$-$1026$   & $1,317,384$     & $4.41$       & $9.98$           & $6.38$       & $\bf{13.29}$     & $7.53$       & $14.95$  \\
 DNN-concat            & $3$      & $1539$-$513$-$775$-$1539$-$1026$ & $3,953,953$     & $4.57$       & $10.07$          & $\bf{6.64}$  & $13.21$          & $7.74$       & $15.21$  \\
 VPNN~\cite{fan17dlm}  & $3$      & $513$-$513$-$513$-$513$-$1026$   & $3,952,152$     & $4.59$       & $10.05$          & $6.57$       & $12.78$          & $7.85$       & $15.41$  \\
 ABIPNN                & $3$      & $513$-$513$-$513$-$513$-$1026$   & $3,952,152$     & $\bf{4.61}$  & $\bf{10.10}$     & $6.61$       & $13.08$          & $\bf{7.86}$  & $\bf{15.67}$  \\
 \hline
\end{tabular}
\end{table*}

A na\"ive DNN approach for source separation, referred to as \emph{DNN-simple} below, takes the spectrum of each frame of the mixture as input, and estimates the spectra of that frame for the vocal and the accompaniment parts respectively.  This is done frame-by-frame, finally leading to the estimated (recovered) spectrograms  $\mathbf{\tilde{Y}_1}$ and $\mathbf{\tilde{Y}_2}$ of the two sources. 
Then, we use the Weiner filter to compute the following soft time-frequency mask to smooth the source separation results. 
\begin{equation}
\mathbf{m}(f)=\frac{\vert\mathbf{\tilde{Y}_1}(f)\vert}{\vert\mathbf{\tilde{Y}_1}(f)\vert+\vert\mathbf{\tilde{Y}_2}(f)\vert} \,,
\end{equation}
where $f=1,2, ..., \mathbf{\textit{F}}$ denotes different frequency bins. The estimated spectra $\mathbf{\tilde{s}_1}$ and $\mathbf{\tilde{s}_2}$, corresponding to vocals and accompaniments, respectively, are produced by:
\begin{equation}
\begin{split} 
&\mathbf{\tilde{s}_1}(f)=\mathbf{m}(f)\mathbf{z}(f)\textrm{,} \\
&\mathbf{\tilde{s}_2}(f)=(1-\mathbf{m}(f))\mathbf{z}(f)\textrm{,}
\end{split}
\end{equation}
where $\mathbf{z}(f)$ is the magnitude spectra of the input mixture.  The estimated spectra $\mathbf{\tilde{s}_1}$ and $\mathbf{\tilde{s}_2}$ are finally transformed back to the time-domain by the inverse short time Fourier transform (ISTFT), assuming that the mixture and the two sources share the same phase. 
Parameters of the NN are learned by using the MSE between the estimated spectra and the groundtruth source spectra.



We can improve the performance of DNN-simple by adding the \emph{temporal context} of each frame to the input \cite{zhang16taslp}.  Specifically, in addition to the current frame, we add the spectra of the previous-$f$ and subsequent-$f$ frames to compose a real-valued matrix.  For a conventional NN, we can take the vectorized version of the matrix (which amounts to concatenating the spectra of these $2f+1$ frames) as the input. We refer to this method as \emph{DNN-concat}.  
Alternatively, we can view the $2f+1$ frames as different dimensions and use a vector NN to model the interaction between different frames. Please see Fig.~\ref{fig:singing_voice_separation_arch} for an illustration.

Because ABIPNN can deal with input of any dimensions, in principle $f$ can take any values. However, in this experiment we intend to compare the performance of ABIPNN with an existing vector NN architecture. We therefore set $f=1$ (i.e., considering only the two neighboring frames) so that we can compare ABIPNN with the classic vector product neural network (VPNN) \cite{nitta93ijcnn}, which can deal with only $N=3$.
The major difference between ABIPNN (with $N=3$) and VPNN is that the former uses circular convolutions.



\begin{figure}
\begin{center}
(a) 009 - Bobby Nobody - Stitch Up \\
\vspace{2mm}
\includegraphics[trim=1.8cm 0.3cm 1.5cm 0.3cm, clip, scale=0.32]{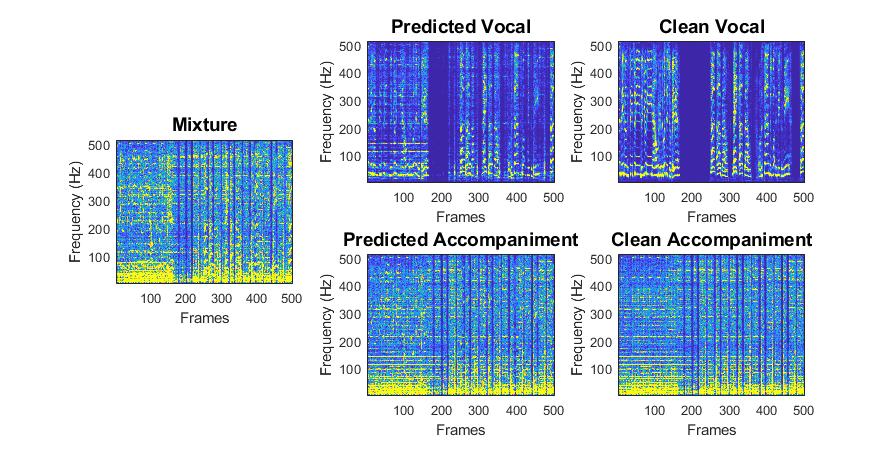} \\
\vspace{2mm}
(b) 039 - Swinging Steaks - Lost My Way \\
\vspace{2mm}
\includegraphics[trim=1.8cm 0.3cm 1.5cm 0.3cm, clip, scale=0.32]{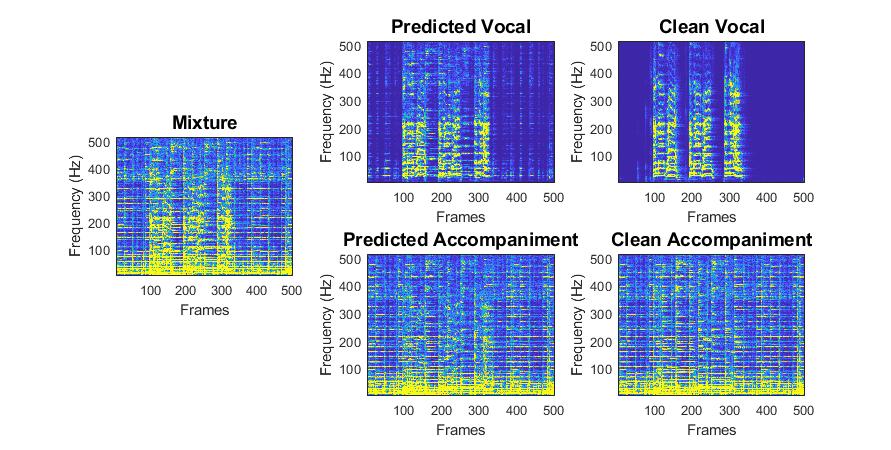}
\caption{Examples of singing voice separation employing ABIPNN on the test part of the DSD100 dataset.}
\label{fig:svs_examples}
\end{center}
\end{figure}

We use in our experiments the Demixing Secret Database (DSD100), which was used in the Signal Separation Evaluation Campaign (SiSEC) in 2016 \cite{sisec16mus,liutkus17springer}. 
It is made up of $100$ full-track professionally-produced music recordings of different styles. It can be used for training and testing for source separation algorithms, because it includes both the stereophonic mixtures and the original stereo sources. The database is divided into a development set and a test set by the organizers of SiSEC 2016, each consisting of $50$ songs. The duration of the songs ranges from 2'22'' to 7'20'', and the average duration is 4'10''. All the songs are sampled at a sampling rate of $44\,100$ Hz. 



To reduce computational cost, all songs are downsampled to $8000$ Hz. For each song, we compute STFT with a $1024$-point window and a $256$-point hop size. 
Then we map each t-f unit of the magnitude spectrum to a three-dimensional vector to serve as the input to ABIPNN. Among the training frames, 10\% of the tensor frames are selected as validation data. The training epoch is set to $1000$ and the training process is stopped if the MSE of the validation set does not decrease for $20$ epochs. The performance is measured in terms of source to distortion ratio (SDR), source to interferences ratio (SIR), and source to artifact ratio (SAR), as calculated by the Blind Source Separation (BSS) Eval Toolbox v3.0 \cite{vincent06taslp}. 



\begin{figure}
\begin{center}
\includegraphics[trim=0.3cm 0.1cm 0.5cm 0.1cm, clip, scale=0.65]{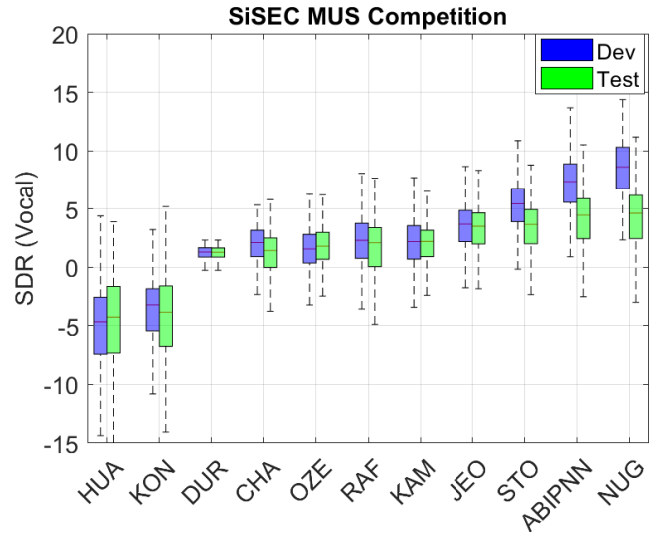}
\caption{Vocal results (in SDR) of DSD100 dataset, sorted by median values of the test part of each submission.
}
\label{fig:sisec_competition_result}
\end{center}
\end{figure}

Table~\ref{tab:dsd100_result_with_nn} shows some details of the evaluated methods and their results. We use the same topology $513$-$513$-$513$-$513$-$1026$ for most methods, except for DNN-concat. We use more neurons for DNN-concat so that it has the same total number of parameters as ABIPNN and VPNN.


In Table~\ref{tab:dsd100_result_with_nn}, vector NNs indeed outperforms conventional DNN methods, demonstrating the effectiveness of considering the interactions between frames. Although DNN-concat has more parameters than DNN-simple, there is little performance difference between the two, showing that the DNN architecture cannot nicely take advantage of the information provided by the temporal context. ABIPNN does not outperform VPNN in all the three metrics, so it is hard to say ABIPNN is better than VPNN in this evaluation. However, we note that it is possible to further improve the result of ABIPNN by using larger $N$, but this is not the case for VPNN.



Fig.~\ref{fig:svs_examples} shows the spectrograms of the input mixture, separation results by ABIPNN, and the original sources for two songs randomly picked from the test set of DSD100. We can see that the separation results (marked as ``predicted vocal'' or ``predicted accompaniment'') resemble the original sources.

Finally, in Fig.~\ref{fig:sisec_competition_result}, we compare the median SDR of the vocal part of ABIPNN with the methods that have been evaluated for SiSEC 2016~\cite{liutkus17springer}, including 
NUG~\cite{nugraha16taslp}, STO~\cite{stoter16icassp}, OZE~\cite{ozerov12taslp}, KON~\cite{huang15taslp}, KAM~\cite{liutkus15icassp}, JEO~\cite{jeong17springer}, HUA~\cite{huang12icassp}, 
DUR\cite{durrieu11jstsp}, CHA\cite{chandna17lvass}, and RAF \cite{raffii13taslp}. 
Among them, CHA is based on CNN, and KON, STO, NUG are based on DNNs. For a fair comparison, we only select those submissions that are not trained with augmentation data. 
Moreover, we show the vocal SDR here instead of other metrics, for saving space (following \cite{liutkus17springer}) 
and for SDR is usually considered as more important than the other metrics. 
It is also a convention in SiSEC to show the result for both the development and the test sets.
From Fig.~\ref{fig:sisec_competition_result}, we see  that the performance of ABIPNN is comparable to NUG~\cite{nugraha16taslp}, which uses expectation maximization and DNNs. 

 

\section{Conclusion}
\label{sec:conclusion}
This paper proposes a novel vector-valued neuron which employs arbitrary bilinear products for feedforward and backpropagation. The proposed architecture generalizes and extends all existing vector-valued neurons and is useful for datasets where each training sample is a multidimensional vector. Through bilinear products, the vector neural network captures the associations among different entries in the same position in each vector. The model can be trained efficiently by using vector error backpropagation through the Adam algorithm. Experimental results on multispectral denoising and singing voice separation show that our proposed model performs better than conventional NNs. Future work involves three directions. First, we will apply the technique of vector neural learning with bilinear product on convolutional neural network (CNN) to deal with spatial data such as images. Next, we will compare the performance of vector neural learning on CNN to other deep models with classification or regression tasks. Finally, we observe that the training speed of the proposed network grows quadratically as a function of the dimensionality $N$. Future work has to be done to make the training more scalable to higher values of $N$.



\appendices
\section{Bilinear Products for ABIPNN}
\label{appendix:bilinear_product_induction}
In Section~\ref{subsec:circular_convolution_induction}, we use circular convolution as the bilinear product in ABIPNN and briefly derive the results of matrices $\left[\mathbf{a}_{j}^{l-1}\right]_{\bullet}^{\dagger}$ and $\left[\mathbf{w}_{ki}^{l+1}\right]_{\bullet}$ used in the backpropagation process. In this section, we will derive some more bilinear products in detail. The feedforward connection is still described between layer $l-1$ and layer $l$.

\subsection{Vector Product}
Here $\mathbf{w}_{ij}^l\bullet\mathbf{a}_{j}^{l-1}$ stands for the usual vector product, $\mathbf{w}_{ij}^l=\left[w_{ij1}^l\textrm{,}~w_{ij2}^l\textrm{,}~w_{ij3}^l\right]^T\in \mathbb{R}^3$, and the input $\mathbf{a}_{j}^{l-1}=\left[a_{j1}^{l-1}\textrm{,}~a_{j2}^{l-1}\textrm{,}~a_{j3}^{l-1}\right]^T\in \mathbb{R}^3$ are all three-dimensional vectors $(N=3)$. The matrix representation of $\mathbf{w}_{ij}^l\bullet\mathbf{a}_{j}^{l-1}$ is:
\renewcommand{\arraystretch}{1.5}
\begin{equation}
\begin{split}
\left[\mathbf{w}_{ij}^{l}\right]_{\bullet}=
\begin{bmatrix}
0             & -w_{ij3}^{l}    & w_{ij2}^{l}   \\ 
w_{ij3}^{l}   & 0               & -w_{ij1}^{l}  \\
-w_{ij2}^{l}  & w_{ij1}^{l}     & 0             \\
\end{bmatrix} \textrm{,}
\end{split}
\end{equation}
where the weight vector is formulated as a $3 \times 3$ square matrix with zeros on the main diagonal. When calculating $\frac{\partial C}{\partial \mathbf{w}_{ij}^l}$,
the matrix $\left[\mathbf{a}_{j}^{l-1}\right]_{\bullet}^{\dagger}$ in 
Eq.~\ref{eq:partial_C_over_partial_wij_result} can be extended as follows:
\renewcommand{\arraystretch}{1.5}
\begin{equation}
\begin{split}
\left[\mathbf{a}_{j}^{l-1}\right]_{\bullet}^{\dagger}
=
\begin{bmatrix}
& 0            & a_{j3}^{l-1} &-a_{j2}^{l-1} \\
&-a_{j3}^{l-1} & 0            & a_{j1}^{l-1} \\
& a_{j2}^{l-1} &-a_{j1}^{l-1} & 0            \\
\end{bmatrix} \textrm{.} 
\end{split}
\end{equation}
Then, we calculate $\frac{\partial \mathbf{z}_{k}^{l+1}}{\partial \mathbf{a}_{i}^l}$ in Eq.~\ref{eq:partial_zk_over_partial_ai_result} using the following matrix:
\begin{equation}
\begin{split}
\left[\mathbf{w}_{ki}^{l+1}\right]_{\bullet}
=
\begin{bmatrix}
& 0             &-w_{ki3}^{l+1} & w_{ki2}^{l+1} \\
& w_{ki3}^{l+1} & 0             &-w_{ki1}^{l+1} \\
&-w_{ki2}^{l+1} & w_{ki1}^{l+1} & 0            \\
\end{bmatrix} \textrm{.}
\end{split}
\end{equation}

ABIPNN with this product is identical to the three-dimensional vector product neural network \cite{nitta93ijcnn}.

\subsection{Quaternion Multiplication}
Quaternions are four-dimensional numbers $a+bi+cj+dk$ with the multiplication rule $i^2=j^2=k^2=ijk=-1$. Here $\mathbf{w}_{ij}^l\bullet \mathbf{a}_{j}^{l-1}$ stands for quaternion multiplication, the weight $\mathbf{w}_{ij}^l=\left[w_{ij1}^l\textrm{,}~w_{ij2}^l\textrm{,}~w_{ij3}^l\textrm{,}~w_{ij4}^l\right]^T\in \mathbb{R}^4$ and the input $\mathbf{a}_{j}^{l-1}=\left[a_{j1}^{l-1}\textrm{,}~a_{j2}^{l-1}\textrm{,}~a_{j3}^{l-1}\textrm{,}~a_{j4}^{l-1}\right]^T\in \mathbb{R}^4$ are all four-dimensional vectors $(N=4)$. The matrix representation of $\mathbf{w}_{ij}^l\bullet \mathbf{a}_{j}^{l-1}$ is:
\renewcommand{\arraystretch}{1.5}
\begin{equation}
\begin{split}
\left[\mathbf{w}_{ij}^{l}\right]_{\bullet}=
\begin{bmatrix}
w_{ij1}^{l}   &-w_{ij2}^{l}     &-w_{ij3}^{l}   &-w_{ij4}^{l}  \\ 
w_{ij2}^{l}   & w_{ij1}^{l}     &-w_{ij4}^{l}   & w_{ij3}^{l}  \\
w_{ij3}^{l}   & w_{ij4}^{l}     & w_{ij1}^{l}   &-w_{ij2}^{l}  \\
w_{ij4}^{l}   &-w_{ij3}^{l}     & w_{ij2}^{l}   & w_{ij1}^{l}  \\
\end{bmatrix} \textrm{,}
\end{split}
\end{equation}
where the weight vector is formulated as a $4 \times 4$ square matrix with $w_{ij1}^l$ on the main diagonal. The matrix $\left[\mathbf{a}_{j}^{l-1}\right]_{\bullet}^{\dagger}$ for the calculation of $\frac{\partial C}{\partial \mathbf{w}_{ij}^l}$ in Eq.~\ref{eq:partial_C_over_partial_wij_result} can be extended as follows:
\begin{equation}
\begin{split}
\left[\mathbf{a}_{j}^{l-1}\right]_{\bullet}^{\dagger}
=
\begin{bmatrix}
a_{j1}^{l-1}   &-a_{j2}^{l-1}     &-a_{j3}^{l-1}   &-a_{j4}^{l-1}  \\ 
a_{j2}^{l-1}   & a_{j1}^{l-1}     & a_{j4}^{l-1}   &-a_{j3}^{l-1}  \\
a_{j3}^{l-1}   &-a_{j4}^{l-1}     & a_{j1}^{l-1}   & a_{j2}^{l-1}  \\
a_{j4}^{l-1}   & a_{j3}^{l-1}     &-a_{j2}^{l-1}   & a_{j1}^{l-1}  \\
\end{bmatrix} \textrm{.}
\end{split}
\end{equation}
After that, we can use the following matrix $\left[\mathbf{w}_{ki}^{l+1}\right]_{\bullet}$ when computing $\frac{\partial \mathbf{z}_{k}^{l+1}}{\partial \mathbf{a}_{i}^l}$ in Eq.~\ref{eq:partial_zk_over_partial_ai_result}:
\renewcommand{\arraystretch}{1.5}
\begin{equation}
\begin{split}
\left[\mathbf{w}_{ki}^{l+1}\right]_{\bullet}=
\begin{bmatrix}
w_{ki1}^{l+1}   &-w_{ki2}^{l+1}     &-w_{ki3}^{l+1}   &-w_{ki4}^{l+1}  \\ 
w_{ki2}^{l+1}   & w_{ki1}^{l+1}     &-w_{ki4}^{l+1}   & w_{ki3}^{l+1}  \\
w_{ki3}^{l+1}   & w_{ki4}^{l+1}     & w_{ki1}^{l+1}   &-w_{ki2}^{l+1}  \\
w_{ki4}^{l+1}   &-w_{ki3}^{l+1}     & w_{ki2}^{l+1}   & w_{ki1}^{l+1}  \\
\end{bmatrix} \textrm{.}
\end{split}
\end{equation}

With quaternion multiplication, the ABIPNN is equivalent to the quaternion-valued neural network \cite{arena94iscas}.

\subsection{Seven-Dimensional Vector Product}
The seven-dimensional vector product is defined as per \cite{lounesto01}. Here $\mathbf{w}_{ij}^{l}\bullet\mathbf{a}_{j}^{l-1}$ stands for the seven-dimensional vector product, $\mathbf{w}_{ij}^l=\left[w_{ij1}^l\textrm{,}~w_{ij2}^l\textrm{,}\dots\textrm{,}~w_{ij7}^l\right]^T\in \mathbb{R}^7$ and $\mathbf{a}_{j}^{l-1}=\left[a_{j1}^{l-1}\textrm{,}~a_{j2}^{l-1}\textrm{,}\dots\textrm{,}~a_{j7}^{l-1}\right]^T\in \mathbb{R}^7$ are both seven-dimensional vectors $(N=7)$, and the matrix representation of $\mathbf{w}_{ij}^{l}\bullet\mathbf{a}_{j}^{l}$ is:
\renewcommand{\arraystretch}{1.5}
\begin{equation}
\setlength{\arraycolsep}{3pt}
\begin{split}
&\left[\mathbf{w}_{ij}^{l}\right]_{\bullet}=\\
&\begin{bmatrix}
 0    		   &-w_{ij4}^{l}    &-w_{ij7}^{l}   & w_{ij2}^{l}   &-w_{ij6}^{l}  & w_{ij5}^{l}   & w_{ij3}^{l} \\ 
 w_{ij4}^{l}   & 0     			&-w_{ij5}^{l}   &-w_{ij1}^{l}   & w_{ij3}^{l}  &-w_{ij7}^{l}   & w_{ij6}^{l} \\
 w_{ij7}^{l}   & w_{ij5}^{l}    & 0   			&-w_{ij6}^{l}   &-w_{ij2}^{l}  & w_{ij4}^{l}   &-w_{ij1}^{l} \\
-w_{ij2}^{l}   & w_{ij1}^{l}    & w_{ij6}^{l}   & 0			    &-w_{ij7}^{l}  &-w_{ij3}^{l}   & w_{ij5}^{l} \\
 w_{ij6}^{l}   &-w_{ij3}^{l}    & w_{ij2}^{l}   & w_{ij7}^{l}   & 0   		   &-w_{ij1}^{l}   &-w_{ij4}^{l} \\ 
-w_{ij5}^{l}   & w_{ij7}^{l}    &-w_{ij4}^{l}   & w_{ij3}^{l}   & w_{ij1}^{l}  & 0   		   &-w_{ij2}^{l} \\ 
-w_{ij3}^{l}   &-w_{ij6}^{l}    & w_{ij1}^{l}   &-w_{ij5}^{l}   & w_{ij4}^{l}  & w_{ij2}^{l}   & 0           \\ 
\end{bmatrix}\textrm{,}
\end{split}
\end{equation}
where the weight vector is formulated as a $7 \times 7$ square matrix in a similar way as the vector product one. The matrix $\left[\mathbf{a}_{j}^{l-1}\right]_{\bullet}^{\dagger}$ in Eq.~\ref{eq:partial_C_over_partial_wij_result} becomes:
\begin{equation}
\setlength{\arraycolsep}{3pt}
\begin{split}
&\left[\mathbf{a}_{j}^{l-1}\right]_{\bullet}^{\dagger}=\\
&\begin{bmatrix}
 0    		    & a_{j4}^{l-1}    & a_{j7}^{l-1}   &-a_{j2}^{l-1}   & a_{j6}^{l-1}  &-a_{j5}^{l-1}   &-a_{j3}^{l-1} \\ 
-a_{j4}^{l-1}   & 0     		  & a_{i5}^{l-1}   & a_{j1}^{l-1}   &-a_{j3}^{l-1}  & a_{j7}^{l-1}   &-a_{j6}^{l-1} \\
-a_{j7}^{l-1}   &-a_{j5}^{l-1}    & 0   	       & a_{j6}^{l-1}   & a_{j2}^{l-1}  &-a_{j4}^{l-1}   & a_{j1}^{l-1} \\
 a_{j2}^{l-1}   &-a_{j1}^{l-1}    &-a_{j6}^{l-1}   & 0			    & a_{j7}^{l-1}  & a_{j3}^{l-1}   &-a_{j5}^{l-1} \\
-a_{j6}^{l-1}   & a_{j3}^{l-1}    &-a_{j2}^{l-1}   &-a_{j7}^{l-1}   & 0   		    & a_{j1}^{l-1}   & a_{j4}^{l-1} \\ 
 a_{j5}^{l-1}   &-a_{j7}^{l-1}    & a_{j4}^{l-1}   &-a_{j3}^{l-1}   &-a_{j1}^{l-1}  & 0   		     & a_{j2}^{l-1} \\ 
 a_{j3}^{l-1}   & a_{j6}^{l-1}    &-a_{j1}^{l-1}   & a_{j5}^{l-1}   &-a_{j4}^{l-1}  &-a_{j2}^{l-1}   & 0            \\ 
\end{bmatrix} \textrm{,}
\end{split}
\end{equation}
where the signs are flipped when compared to the matrix $\left[\mathbf{w}_{ij}^{l}\right]_{\bullet}$ above. The matrix $\left[\mathbf{w}_{ki}^{l+1}\right]_{\bullet}$ for the seven-dimensional vector product is then extended as:
\begin{equation}
\setlength{\arraycolsep}{3pt}
\begin{split}
&\left[\mathbf{w}_{ki}^{l+1}\right]_{\bullet}=\\
&\begin{bmatrix}
 0    		   &-w_{ki4}^{l+1}  &-w_{ki7}^{l+1} & w_{ki2}^{l+1} &-w_{ki6}^{l+1}  & w_{ki5}^{l+1} & w_{ki3}^{l+1} \\ 
 w_{ki4}^{l+1} & 0     			&-w_{ki5}^{l+1} &-w_{ki1}^{l+1} & w_{ki3}^{l+1}  &-w_{ki7}^{l+1} & w_{ki6}^{l+1} \\
 w_{ki7}^{l+1} & w_{ki5}^{l+1}  & 0   			&-w_{ki6}^{l+1} &-w_{ki2}^{l+1}  & w_{ki4}^{l+1} &-w_{ki1}^{l+1} \\
-w_{ki2}^{l+1} & w_{ki1}^{l+1}  & w_{ki6}^{l+1} & 0			    &-w_{ki7}^{l+1}  &-w_{ki3}^{l+1} & w_{ki5}^{l+1} \\
 w_{ki6}^{l+1} &-w_{ki3}^{l+1}  & w_{ki2}^{l+1} & w_{ki7}^{l+1} & 0   		     &-w_{ki1}^{l+1} &-w_{ki4}^{l+1} \\ 
-w_{ki5}^{l+1} & w_{ki7}^{l+1}  &-w_{ki4}^{l+1} & w_{ki3}^{l+1} & w_{ki1}^{l+1}  & 0   		     &-w_{ki2}^{l+1} \\ 
-w_{ki3}^{l+1} &-w_{ki6}^{l+1}  & w_{ki1}^{l+1} &-w_{ki5}^{l+1} & w_{ki4}^{l+1}  & w_{ki2}^{l+1} & 0           \\ 
\end{bmatrix}\textrm{.}
\end{split}
\end{equation}

\subsection{Skew Circular Convolution}
The skew-circular convolution is obtained by replacing the circulant matrix in Eq.~\ref{eq:circular_convolution} by a skew-circulant one \cite{davis12circulant}. The matrix representation of $\mathbf{w}_{ij}^{l}\bullet\mathbf{a}_{j}^{l-1}$ then becomes:
\renewcommand{\arraystretch}{1.5}
\begin{equation}
\label{eq:skew_circular_convolution}
\begin{split}
\left[\mathbf{w}_{ij}^{l}\right]_{\bullet}=
\begin{bmatrix}
w_{ij1}^l  &-w_{ijN}^l      &-w_{ij(N-1)}^l  & \dots     &-w_{ij2}^l   \\ 
w_{ij2}^l  & w_{ij1}^l      &-w_{ijN}^l      & \dots     &-w_{ij3}^l   \\
w_{ij3}^l  & w_{ij2}^l      & w_{ij1}^l      & \dots     &-w_{ij4}^l   \\
\vdots     & \vdots         & \vdots         & \ddots    & \vdots      \\
w_{ijN}^l  & w_{ij(N-1)}^l  & w_{ij(N-2)}^l  & \dots     & w_{ij1}^l   \\
\end{bmatrix}\textrm{,}
\end{split}
\end{equation}
where the weight vector is formulated as an $N \times N$ square matrix with $w_{ij1}^l$ on the main diagonal and the upper triangular part of the weight matrix is multiplied by minus one. 
When we compute $\frac{\partial C}{\partial \mathbf{w}_{ij}^l}$ in Eq.~\ref{eq:partial_C_over_partial_wij_result}, $\left[\mathbf{a}_{j}^{l-1}\right]_{\bullet}^{\dagger}$ is extended as follows:
\begin{equation}
\begin{split}
\left[\mathbf{a}_{j}^{l-1}\right]_{\bullet}^{\dagger}=
\begin{bmatrix}
a_{j1}^{l-1}  &-a_{jN}^{l-1}      &-a_{j(N-1)}^{l-1}  & \dots     &-a_{j2}^{l-1}  \\ 
a_{j2}^{l-1}  & a_{j1}^{l-1}      &-a_{jN}^{l-1}      & \dots     &-a_{j3}^{l-1}  \\
a_{j3}^{l-1}  & a_{j2}^{l-1}      & a_{j1}^{l-1}      & \dots     &-a_{j4}^{l-1}  \\
\vdots        & \vdots            & \vdots            & \ddots    & \vdots        \\
a_{jN}^{l-1}  & a_{j(N-1)}^{l-1}  & a_{j(N-2)}^{l-1}  & \dots     & a_{j1}^{l-1}  \\
\end{bmatrix} \textrm{.}
\end{split}
\end{equation}
When $\frac{\partial \mathbf{z}_{k}^{l+1}}{\partial \mathbf{a}_{i}^l}$ is computed, the matrix $\left[\mathbf{w}_{ki}^{l+1}\right]_{\bullet}$ in Eq.~\ref{eq:partial_zk_over_partial_ai_result} is extended for skew circular convolution as follows:
\begin{equation}
\begin{split}
\left[\mathbf{w}_{ki}^{l+1}\right]_{\bullet}=
\begin{bmatrix}
w_{ki1}^{l+1}  &-w_{kiN}^{l+1}      &-w_{ki(N-1)}^{l+1}  & \dots     &-w_{ki2}^{l+1}   \\ 
w_{ki2}^{l+1}  & w_{ki1}^{l+1}      &-w_{kiN}^{l+1}      & \dots     &-w_{ki3}^{l+1}   \\
w_{ki3}^{l+1}  & w_{ki2}^{l+1}      & w_{ki1}^{l+1}      & \dots     &-w_{ki4}^{l+1}   \\
\vdots         & \vdots             & \vdots             & \ddots    & \vdots          \\
w_{kiN}^{l+1}  & w_{ki(N-1)}^{l+1}  & w_{ki(N-2)}^{l+1}  & \dots     &-w_{ki1}^{l+1}   \\
\end{bmatrix} \textrm{.}
\end{split}
\end{equation}
If $N=2$, then ABIPNN becomes a complex-valued neural network \cite{nitta97nn}. For even $N\geq 2$, the skew circular convolution is also known as planar complex multiplication \cite{olariu02elsevier}.

\subsection{Reverse-Time Circular Convolution}
The reversed-time circular convolution is obtained by flipping the circulant matrix upside down (see also \cite{powell90iscs}). The matrix representation of $\mathbf{w}_{ij}^{l}\bullet\mathbf{a}_{j}^{l-1}$ becomes:
\renewcommand{\arraystretch}{1.5}
\begin{equation}
\label{eq:reverse_time_circular_convolution}
\begin{split}
&\left[\mathbf{w}_{ij}^{l}\right]_{\bullet}= \\
&\begin{bmatrix}
w_{ijN}^l      & w_{ij(N-1)}^l    & w_{ij(N-2)}^l   & \dots     & w_{ij1}^l       \\ 
w_{ij(N-1)}^l  & w_{ij(N-2)}^l    & w_{ij(N-3)}^l   & \dots     & w_{ijN}^l       \\
w_{ij(N-2)}^l  & w_{ij(N-3)}^l    & w_{ij(N-4)}^l   & \dots     & w_{ij(N-1)}^l   \\
\vdots         & \vdots           & \vdots          & \ddots    & \vdots          \\
w_{ij1}^l  	   & w_{ijN}^l      & w_{ij(N-1)}^l   & \dots     & w_{ij2}^l       \\
\end{bmatrix}\textrm{,}
\end{split}
\end{equation}
where the weight vector is formulated as an $N \times N$ square matrix which is the same as the weight matrix for circular convolution rotated by 90 degrees. The matrix $\left[\mathbf{a}_{j}^{l-1}\right]_{\bullet}^{\dagger}$ in Eq.~\ref{eq:partial_C_over_partial_wij_result} is extended as follows:
\begin{equation}
\begin{split}
&\left[\mathbf{a}_{j}^{l-1}\right]_{\bullet}^{\dagger}= \\
&\begin{bmatrix}
a_{jN}^{l-1}      & a_{j(N-1)}^{l-1}  & a_{j(N-2)}^{l-1}  & \dots     & a_{j1}^{l-1}      \\ 
a_{j(N-1)}^{l-1}  & a_{j(N-2)}^{l-1}  & a_{j(N-3)}^{l-1}  & \dots     & a_{jN}^{l-1}      \\
a_{j(N-2)}^{l-1}  & a_{2(N-3)}^{l-1}  & a_{j(N-4)}^{l-1}  & \dots     & a_{j(N-1)}^{l-1}  \\
\vdots            & \vdots            & \vdots            & \ddots    & \vdots            \\
a_{j1}^{l-1}      & a_{jN}^{l-1}      & a_{j(N-1)}^{l-1}  & \dots     & a_{j2}^{l-1}      \\
\end{bmatrix} \textrm{,}
\end{split}
\end{equation}
Then, the derivative $\frac{\partial \mathbf{z}_{k}^{l+1}}{\partial \mathbf{a}_{i}^l}$ in Eq.~\ref{eq:partial_zk_over_partial_ai_result} is extended for reverse-time circular convolution by the following:
\begin{equation}
\begin{split}
&\left[\mathbf{w}_{ki}^{l+1}\right]_{\bullet}=\\
&\begin{bmatrix}
w_{kiN}^{l-1}      & w_{ki(N-1)}^{l-1}  & w_{ki(N-2)}^{l-1}  & \dots     & w_{ki1}^{l-1}      \\ 
w_{ki(N-1)}^{l-1}  & w_{ki(N-2)}^{l-1}  & w_{ki(N-3)}^{l-1}  & \dots     & w_{kiN}^{l-1}      \\
w_{ki(N-2)}^{l-1}  & w_{ki(N-3)}^{l-1}  & w_{ki(N-4)}^{l-1}  & \dots     & w_{ki(N-1)}^{l-1}  \\
\vdots             & \vdots             & \vdots             & \ddots    & \vdots             \\
w_{ki1}^{l-1}      & w_{kiN}^{l-1}      & w_{ki(N-1)}^{l-1}  & \dots     & w_{ki2}^{l-1}      \\
\end{bmatrix}  \textrm{.}
\end{split}
\end{equation}


\ifCLASSOPTIONcaptionsoff
  \newpage
\fi



\bibliographystyle{IEEEtran}
\bibliography{refs.bib}
%


%


\begin{IEEEbiographynophoto}{Zhe-Cheng~Fan} received the M.S. degree in computer science and information engineering from National Taiwan Normal University in 2012. He is currently pursuing the Ph.D. degree in computer science at National Taiwan University. His research interests include machine learning, deep learning, audio signal processing, and singing voice separation.
\end{IEEEbiographynophoto}

\begin{IEEEbiographynophoto}{Tak-Shing~T.~Chan} (M’15)
received the Ph.D. degree in computing from the University of London in 2008. From 2006 to 2008, he was a Scientific Programmer at the University of Sheffield. In 2011, he worked as a Research Associate at the Hong Kong Polytechnic University. He is currently a Postdoctoral Fellow at Academia Sinica. His research interests include signal processing, cognitive informatics, distributed computing, pattern recognition, and hypercomplex analysis.
\end{IEEEbiographynophoto}

\begin{IEEEbiographynophoto}{Yi-Hsuan~Yang} (M'11--SM'17)
is an Associate Research Fellow with Academia Sinica. He received his Ph.D. degree in communication engineering from National Taiwan University in 2010.
He is also a Joint-Appointment Associate Professor with the National Cheng Kung University. His research interests include music information retrieval, affective computing, multimedia, and machine learning. 
Dr. Yang was a recipient of the 2011 IEEE Signal Processing Society Young Author Best Paper Award, the 2012 ACM Multimedia Grand Challenge First Prize, 
and the 2015 Best Conference Paper Award of the IEEE Multimedia Communications Technical Committee. 
He is an author of the book Music Emotion Recognition (CRC Press 2011). 
In 2016, he started his term as an Associate Editor for the \emph{IEEE Transactions on Affective Computing} and the \emph{IEEE Transactions on Multimedia}.
\end{IEEEbiographynophoto}

\begin{IEEEbiographynophoto}{Jyh-Shing~Roger~Jang}(M’93) received the Ph.D. degree in electrical engineering and computer science from the University of California, Berkeley. 
He studied fuzzy logic and artificial neural networks with Prof. L. Zadeh, the father of fuzzy logic. 
He joined MathWorks and has co-authored the Fuzzy Logic Toolbox (for MATLAB). 
He was a Professor with the Department of Computer Science, National Tsing Hua University, from 1995 to 2012. Since 2012, he has been a Professor with the Department of Computer Science and Information Engineering, National Taiwan University. His current research interests include machine learning and pattern recognition, with applications to speech, music, and image processing.
He has over 9000 Google Scholar citations for his seminal paper on adaptive neuro-fuzzy inference systems, published in 1993. 
He has published a book entitled Neuro-Fuzzy and Soft Computing, two books on MATLAB programming, and a book on JavaScript programming. 
\end{IEEEbiographynophoto}
        




\end{document}